\documentclass[8pt,twocolumn]{article}
\usepackage[a4paper, total={7in, 10in}]{geometry}

\usepackage{times}
\usepackage{epsfig}
\usepackage{graphicx}
\usepackage{amsmath}
\usepackage{amssymb}
\usepackage{bbm}
\usepackage{mathtools}
\usepackage{subcaption}
\usepackage{url} 
\usepackage{sidecap}
\usepackage{pifont}
\usepackage{multirow}
\usepackage[table,xcdraw]{xcolor}

\newcommand{\revision}[1]{{#1}}

\newcommand{\cmark}{\textcolor{green}{\ding{51}}}%
\newcommand{\xmark}{\textcolor{red}{\ding{55}}}%
\newcommand*{\rom}[1]{\expandafter\@slowromancap\romannumeral #1@}
\newcommand{\eg}{\textit{e.g.}}
\newcommand{\ie}{\textit{i.e.}}

\newcommand\blfootnote[1]{%
  \begingroup
  \renewcommand\thefootnote{}\footnote{#1}%
  \addtocounter{footnote}{-1}%
  \endgroup
}

\begin{document}








\title{Right for the Right Concept:\\Revising Neuro-Symbolic Concepts by Interacting with their Explanations}

\author{Wolfgang Stammer$^{1,}$\thanks{Equal contribution}
	\and
	Patrick Schramowski$^{1,}$\footnotemark[1]
	\and
	Kristian Kersting$^{1,2}$
	\and
	$^{1}$Computer Science Department, TU Darmstadt, Germany \\
	$^{2}$Centre for Cognitive Science, TU Darmstadt, and Hessian Center for AI (hessian.AI) \\
	{\tt\small \{wolfgang.stammer, schramowski, kersting\}@cs.tu-darmstadt.de}
}

\maketitle

\begin{abstract}
	Most explanation methods in deep learning  map importance estimates for a model's prediction back to the original input space. These ``visual'' explanations are often insufficient, as the model's actual concept remains elusive. Moreover, without insights into the model's semantic concept, it is difficult ---if not impossible--- to intervene on the model's behavior via its explanations, called 
	Explanatory Interactive Learning. Consequently, we propose to intervene on a Neuro-Symbolic scene representation, which allows one to revise the model on the semantic level, e.g. ``never focus on the color to make your decision''. We compiled a novel confounded visual scene data set, the CLEVR-Hans data set, capturing complex compositions of different objects. The results of our experiments on CLEVR-Hans demonstrate that our semantic explanations, i.e. compositional explanations at a per-object level, can identify confounders that are not identifiable using ``visual'' explanations only. More importantly, feedback on this semantic level makes it possible to revise the model from focusing on these factors.
\end{abstract}

\thispagestyle{empty} 

\section{Introduction}


\begin{figure}[t]
	\centering
	\includegraphics[width=0.95\linewidth]{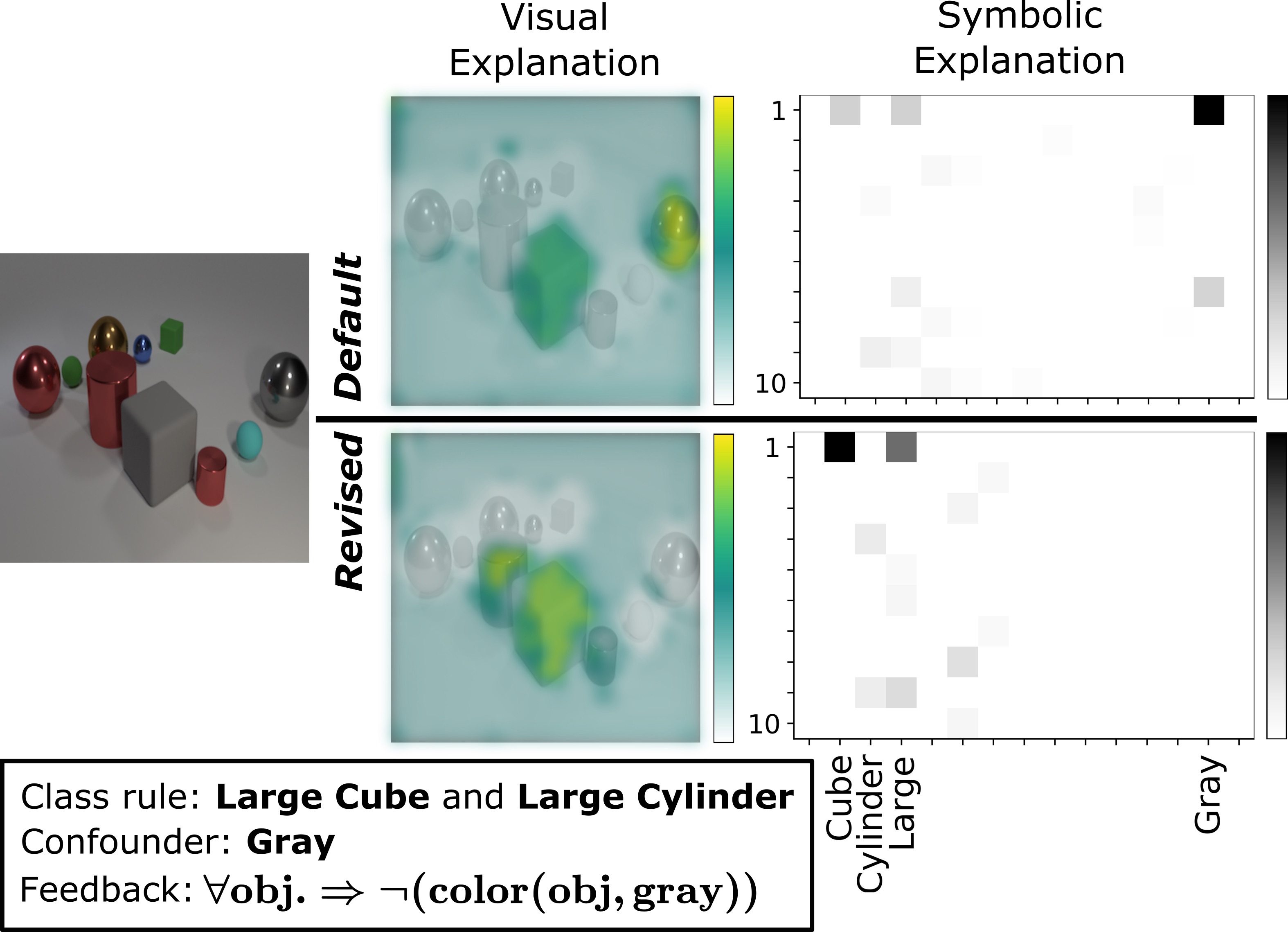}
	\caption{
		\textbf{Neuro-Symbolic explanations are needed to revise deep learning models from focusing on irrelevant features via global feedback rules.}
	}
	\label{fig:example_fig1}
\end{figure}

Machine learning models may show Clever-Hans like moments when solving a task by learning the ``wrong'' thing, \eg making use of confounding factors within a data set. Unfortunately, it is not easy to find out whether, say, a deep neural network is making Clever-Hans-type mistakes because they are not reflected in the standard performance measures such as precision and recall. Instead, one looks at their explanations to see what features the network is actually using~\cite{lapuschkin2019unmasking}. By interacting with the explanations, one may even fix 
Clever-Hans like moments \cite{ross2017right,teso2019explanatory,selvaraju2019taking,schramowski2020making}.
\blfootnote{Published at CVPR 2021.}

This Explanatory Interactive Learning (XIL), however, very much depends on the provided explanations. Most explanation methods in deep learning map importance estimates for a model's prediction back to the original input space 
\cite{selvaraju2017grad, sundararajan2017axiomatic, smilkov2017smoothgrad, Schulz2020Restricting, du2019techniques}. 
This is somewhat reminiscent of a child who points towards something but cannot articulate why something is relevant. In other words, ``visual'' explanations are insufficient if a task requires a concept-level understanding of a model's decision. Without knowledge about and symbolic access to the concept level, it remains difficult---if not impossible---to fix Clever-Hans behavior.

To illustrate this, consider the classification task depicted in Fig.~\ref{fig:example_fig1}. It shows a complex scene consisting of objects, which vary in position, shape, size, material, and color. Now, assume that scenes belonging to the true class show a large cube and a large cylinder. Unfortunately, during training, our deep network only sees scenes with large, gray cubes. Checking the deep model's decision process using visual explanations confirms this: the deep model has learned to largely focus on the gray cube to classify scenes to be positive. An easy fix would be to provide feedback in the form of ``never focus on the color to make your decision'' as it would eliminate the confounding factor. 
Unfortunately, visual explanations do not allow us direct access to the semantic level--- they do not tell us that ``the color gray is an important feature for the task at hand" and we cannot provide feedback at the symbolic level. 

Triggered by this, we present the first Neuro-Symbolic XIL (NeSy XIL) approach that is based on decomposing a visual scene into an object-based, symbolic representation and, in turn, allows one to compute and interact with neuro-symbolic explanations. We demonstrate the advantages of NeSy XIL on a newly compiled, confounded  data set, called CLEVR-Hans. It consists of scenes that can be classified based on specific combinations of object attributes and relations. Importantly, CLEVR-Hans encodes confounders in a way so that the confounding factors are not separable in the original input space, in contrast to many previous confounded computer vision data sets.

To sum up,  this work makes the following contributions:
(i) We confirm empirically on our newly compiled confounded benchmark data set, CLEVR-Hans, 
that Neuro-Symbolic concept learners \cite{mao2019neuro} may show Clever-Hans moments, too. 
(ii) To this end, we devise a novel Neuro-Symbolic concept learner, combining Slot Attention \cite{locatello2020object} and Set Transformer \cite{lee2019set} in an end-to-end differentiable fashion. 
(iii) \revision{We provide a novel loss to revise this Clever-Hans behaviour.}
(iv) Given symbolic annotations about incorrect explanations, even across a set of several instances, we efficiently optimize the Neuro-Symbolic concept learner to be right for better Neuro-Symbolic reasons. 
(v) \revision{Thus we introduce the first XIL approach that works on both the visual and the conceptual level.}
These contributions are important to make progress towards creating conversational explanations between machines and human users 
\cite{weld2019challenge, miller2019explanation}. This is necessary for improved trust development and truly Explanatory Interactive Learning: symbolic abstractions help us, humans, to engage in conversations with one another and to convey our thoughts
efficiently, without the need to specify much detail.\footnote{Source code is available at \url{https://github.com/ml-research/NeSyXIL}}


\section{Related Work on XIL}

Our work touches upon Explainable AI, Explanatory Interactive Learning, and Neuro-Symbolic architectures. 


\textbf{Explainable AI} 
(XAI) methods, in general, are used to evaluate the reasons for a (black-box) model's decision by presenting the model's explanation in a hopefully human-understandable way. Current methods can be divided into various categories based on characteristics \cite{xie2020explainable}, \eg their level of intrinsicality or if they are based on back-propagation computations. 
Across the spectrum of XAI approaches, from backpropagation-based \cite{sundararajan2017axiomatic, ancona2018towards}, to model distillation \cite{ribeiro2016should}, or prototype-based \cite{li2018deep} methods, very often an explanation is created by highlighting or otherwise relating direct input elements to the model's prediction, thus visualizing an explanation at the level of the input space.

Several studies have investigated methods that produce explanations other than these visual explanations, such as multi-modal explanations \cite{huk2018multimodal, wu2019self, rajani2020esprit}, including visual and logic rule explanations \cite{aditya2018explicit, rabold2019enriching}. \cite{mascharka2018transparency, liu2019clevr} investigate methods for creating more interactive explanations, whereas \cite{ciravegnahuman} focuses on creating single-modal, logic-based explanations. Some recent work has also focused on creating concept-based explanations \cite{kimWGCWVS18, zhou2018interpretable, ghorbaniWZK19}.
None of the above studies, however, investigate using the explanations as a means of intervening on the model. 

\textbf{Explanatory interactive learning} (XIL) \cite{ross2017right, selvaraju2019taking, teso2019explanatory, schramowski2020making} merges XAI with an active learning setting. It incorporates XAI in the learning process by involving the human-user \mbox{---interacting} on the \mbox{explanations---} in the training loop. More precisely, the human user can query the model for explanations of individual predictions and respond by correcting the model if necessary, providing a slightly improved \mbox{---but} not necessarily \mbox{optimal---} feedback on the explanations. Thus, as in active learning, the user can provide the correct label if the prediction is wrong. In addition, XIL also allows the user to provide feedback on the explanation. This combination of receiving explanations and user interaction is a strong necessity for gaining trust in the model's behavior \cite{teso2019explanatory, schramowski2020making}. XIL can be applied to differentiable as well as non-differentiable models \cite{schramowski2020making}.

\textbf{Neuro-Symbolic architectures}~\cite{garcez2012neural, yi2018neural, mao2019neuro, hudsonm19,Vedantam2019ProbabilisticNM, garcezGLSST19} make use of data-driven, sub-symbolic representations, and symbol-based reasoning systems. This field of research has received increasing interest in recent years as a means of solving issues of individual subsystems, such as the out-of-distribution generalization problem of many neural networks, by combining the advantages of symbolic and sub-symbolic models. Yi \textit{et al.} \cite{yi2018neural}, for example, propose a Neuro-Symbolic based VQA system based on disentangling visual perception from linguistic reasoning. Each sub-module of their system processes different subtasks, \eg their scene parser decomposes a visual scene into an object-based scene representation. Their reasoning engine then uses this decomposed scene representation rather than directly computing in the original input space. An approach that also relates to the work of Lampert \textit{et al.} \cite{lampert2009learning, lampert2013attribute}.

\section{Motivating Example: Color-MNIST}

\begin{figure}[t]
	\centering
	\includegraphics[width=0.9\linewidth]{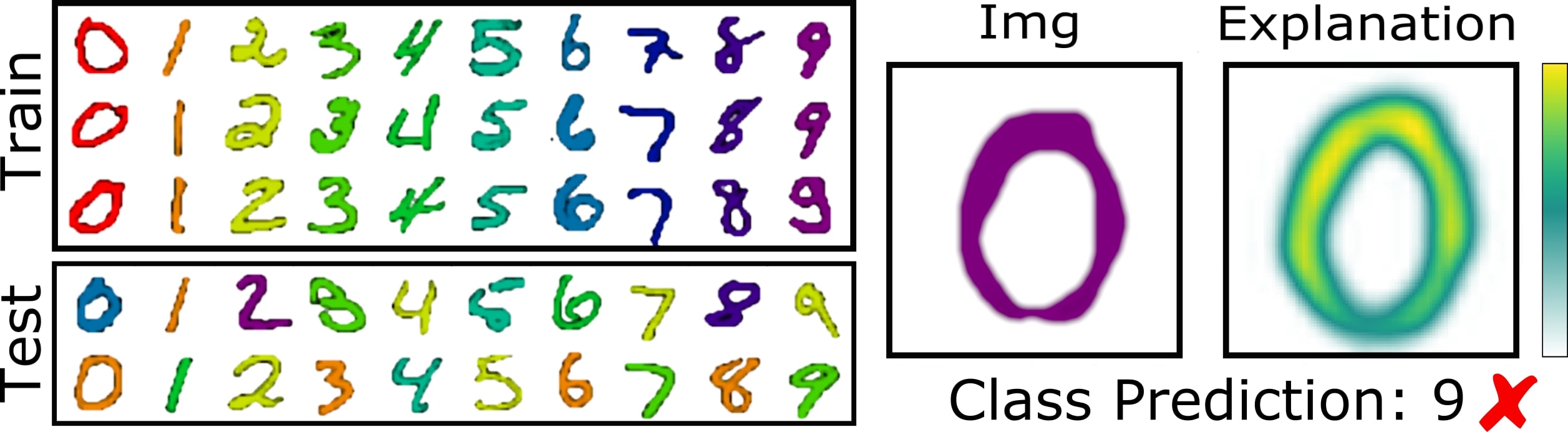}
	\caption{\textbf{Visual Explanations for ColorMNIST.} (Left) the general data distribution between train and test split. 
		(Right) a typical visual explanation of a CNN. Notice digit pixels are considered as important for the wrong prediction.
	}
	\label{fig:color_mnist}
\end{figure}
To illustrate the problem setting, we first revert to a well known confounded toy data set. ColorMNIST \cite{kim2019learning, rieger2020interpretations} consists of colored MNIST digits. Within the training set, each number is confounded with a specific color, whereas in the test set, the color association is shuffled or inverted.

A simple CNN model can reach $100\%$ accuracy on the training set, but only $23\%$ on the test set, indicating that the model has learned to largely focus on the color for accurate prediction rather than the digits themselves. Fig.~\ref{fig:color_mnist} depicts the visual explanation (here created using GradCAM \cite{selvaraju2017grad}) of a zero that is predicted as a nine. Note the zero is colored in the same color as all nines of the training set. 
From the visual explanation it becomes clear that the model is focusing on the correct object, however why the model is predicting the wrong digit label does not become clear without an understanding of the underlying training data distribution.

Importantly, although the model is wrong for the right reason, 
it is a non-trivial problem of interacting with the model to revise its decision using XIL solely based on these explanations. Setting a loss term to correct the explanation (\eg \cite{ross2017right})
on color channels is as non-trivial and inconvenient as un-confounding the data set with counterexamples \cite{teso2019explanatory}. Kim \textit{et al.}~\cite{kim2019learning} describe how to unbias such a data set if the bias is known, using the mutual information between networks trained on separate features of the data set in order for the main network not to focus on bias features. 
Rieger \textit{el al.}~\cite{rieger2020interpretations} propose an explanation penalization loss similar to \cite{ross2017right, selvaraju2019taking, schramowski2020making}, focusing on Contextual Decomposition \cite{murdoch2018beyond} as explanation method. However, the utilized penalization method is task-specific and detached from the model's explanations, resulting in only a slight improvement of a final $31\%$ accuracy (using the inverted ColorMNIST setting).

\section{Neuro-Symbolic Explanatory Interactive Learning}
\begin{figure*}[t]
	\centering
	\includegraphics[width=0.95\textwidth]{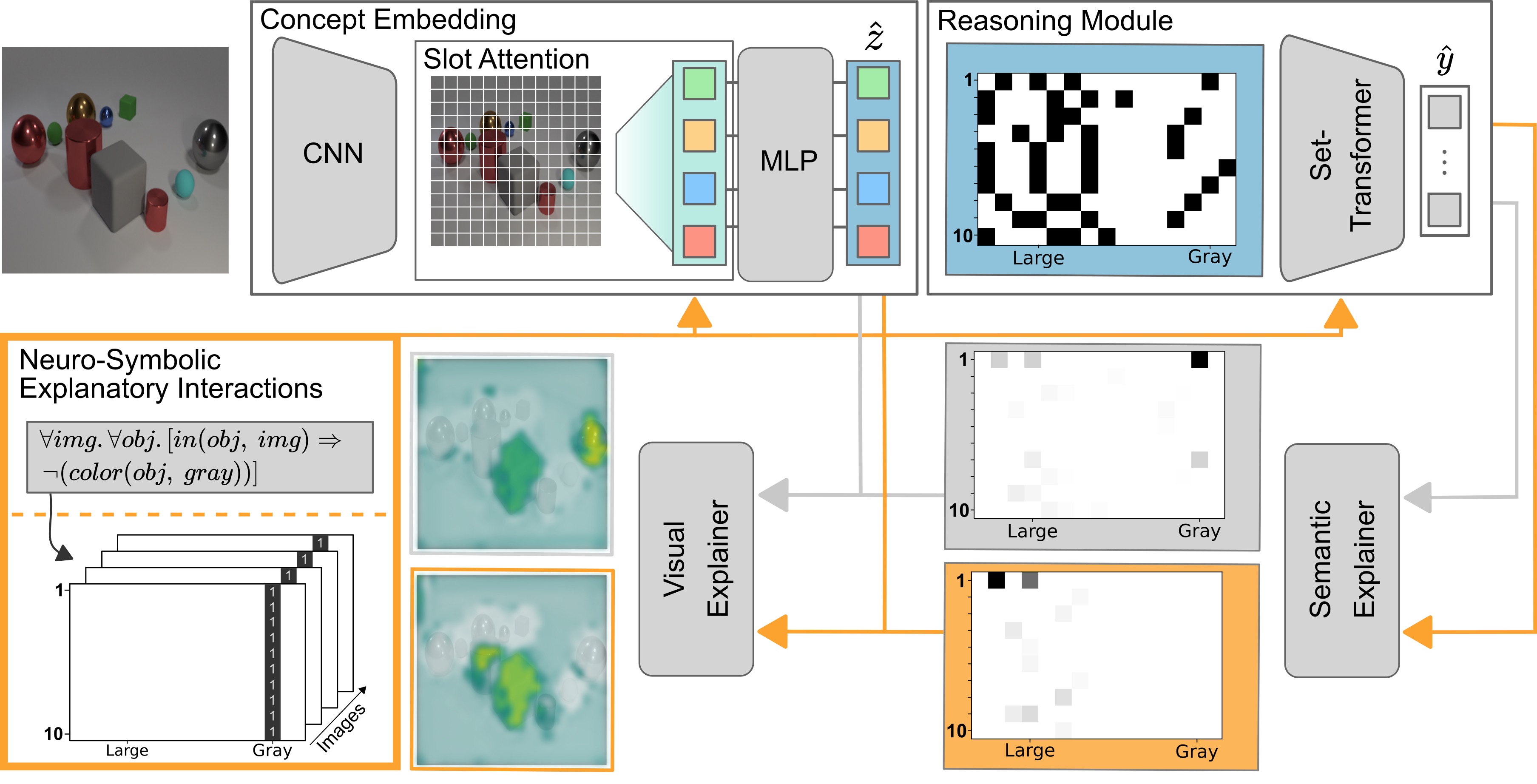}
	\caption{\textbf{Neuro-Symbolic XIL for improved explanations and interaction.} (Top) 
		Neuro-Symbolic Concept Learner with Slot-Attention and Set Transformer. (Bottom) Neuro-Symbolic revision pipeline 
		with explanations of the model before (gray) and after applying the feedback (orange).
	}
	\label{fig:overview}
\end{figure*}
The Color-MNIST example clearly shows that although the input-level explanations of current XAI methods are an important first step towards true explanations of a model's behavior, much ambiguity in a model's decision process 
remains. 
Using XIL on these visual explanations only, it can be difficult to properly intervene on a model. What we require is an understandable, disentangled representation level, which the user can enquire from \mbox{and intervene on.}

\textbf{Neuro-Symbolic Architecture.}
For this purpose, we construct an architecture consisting of two modules, a concept embedding and a reasoning module. 
The concept module's
task is to create a decomposed representation of the input space that can be mapped to human-understandable symbols. 
The task of the reasoning module is to make predictions based on this symbolic representation. 

Fig.~\ref{fig:overview} gives an illustrative overview of our 
approach,
which we formulate more precisely in the following: 
Given an input image $x_i \in X$, whereby \mbox{$X := [x_1, ..., x_N] \in \mathbb{R}^{N \times M}$}, with $X$ being divided into subsets of $N_c$ classes \mbox{$\{X_1, ..., X_{N_c} \} \in X$} and with ground-truth class labels defined as $y \in [0,1]^{N \times N_c}$, we have two modules, the concept embedding module,
$h(x_i) = \hat{z}_i$,
which receives the input sample and encodes it into a symbolic representation, with $\hat{z} \in [0,\ 1]^{N \times D}$. And the reasoning module,
$g(\hat{z}_i) = \hat{y}_i$,
which produces the prediction output, $\hat{y}_i \in [0,1]^{N \times N_c}$, given the symbolic representation.
The exact details of the 
$g(\hat{z}_i)$ and $h(x_i)$ 
depend on the specific implementations of these modules, and will be discussed further in sections below. 

\textbf{Retrieving Neuro-Symbolic Explanations.}
%
Given these two modules, we can extract explanations for the separate tasks, \ie the more general input representation task and the reasoning task. We write an explanation function in a general notation as 
$E(m(\cdot) , \ o, \ s)$,
which retrieves the explanation of a specific module, $m(\cdot)$, given the module's \revision{input $s$, and} the module's output if it is the final module or the explanation of the \revision{following} module if it is not, both summarized as $o$ here. For our approach, we thus have 
$
E^g(g(\cdot), \ \hat{y}_i, \ z_i) =: \hat{e}_{i}^g \label{eq:expl_model1}
$ 
and
$
E^h(h(\cdot), \ \hat{e}_{i}^g, \ x_i) =: \hat{e}_{i}^h \label{eq:expl_model2}
$.
These can represent scalars, vectors, or matrices, depending on the given module and output. $\hat{e}_{i}^g$ represents the explanation of the reasoning module given the final predicted output $\hat{y}_i$, \eg a logic-based rule. $\hat{e}_{i}^h$ presents the explanation of the concept module given the explanation of the reasoning module $\hat{e}_{i}^g$, \eg a visual explanation of a learned concept. In this way, the explanation of the reasoning module is passed back to the concept module in order to receive the explanations of the concept module that contribute to the explanation of the reasoning module. This explanation pass is depicted by the gray arrows of Fig.~\ref{fig:overview}. The exact definition of $E^g$ and $E^h$ used in this work are described below. 

\textbf{Revising Neuro-Symbolic Concepts.}
As we show in our experiments below, also Neuro-Symbolic models are prone to focusing on wrong reasons, \eg confounding factors. In such a case, it is desirable for a user to intervene on the model, \eg via XIL. As errors can result from different modules of the concept learner, the user must create feedback tailored to the individual module that is producing the error. A user thus receives the explanation of a module, \eg $\hat{e}_{i}^g$, and produces an adequate feedback given knowledge of the input sample, $x_i$, the true class label, $y_i$, the model's class prediction $\hat{y}_i$ and possible internal representations, \eg $\hat{z}_i$. 
For the user to interact with the model, the user's feedback must be
mapped back into a representation space of the model. 

In the case of creating feedback for a visual explanation, as in \cite{ross2017right}, \cite{teso2019explanatory} and \cite{schramowski2020making}, the mapping is quite clear: the user gives visual feedback denoting which regions of the input are relevant and which are not. This ``visual'' feedback is transferred to the input space in the form of binary image masks, which we denote as $A^{v}_{i}$. 

The semantic user feedback 
can be in the form of relational functions, $\varphi$, for instance, \textit{``if an image belongs to class $1$ then one object is a large cube''}: 
\begin{center}
	\small
	\noindent\fbox{%
		\parbox{0.9\linewidth}{%
			$\forall img.\ isclass(img, 1) \Rightarrow  \exists obj. [in(obj, img) \land size(obj, large) \land shape(obj, cube)]$ , 
		}%
	}
\end{center}
We define \mbox{$A^s_i := \bigvee_{\varphi} A^{\varphi}_i(\hat{z}_i \models \varphi)$} which describes the disjunction over all relational feedback functions which hold for the symbolic representation, $\hat{z}_i$, of an image, $x_i$.

An important characteristic of the semantic user feedback is that it can describe different levels of generalizability, so that feedback based on a single sample can be transferred to a set of multiple samples. For instance $\varphi$ can hold for an individual sample, all samples of a specific class, $j$, or all samples of the data set. Consequently, the disjunction, $\bigvee_{\varphi}$, can be separated as: 
$
A_{i| y_i = j}^s = A_{i}^{sample} \lor A^{class}_{c=j} \lor A^{all}
$.

For the sake of simplicity, we are not formally introducing relational logic and consider the semantic feedback in tabular form (\textit{cf}. Fig.~\ref{fig:overview}).
To summarize, we have the binary masks for the visual feedback 
$
A^{v}_{i} \in [0,1]^{M}
$
and the semantic feedback
$
A^{s}_{i} \in [0,1]^{D}.
$

For the final interaction we refer to XIL with differentiable models and explanation functions, generally formulated as the explanatory loss term, $ L_{expl} =$
\begin{align}
	\lambda\sum\nolimits_{i=1}^N  r( A_{i}^v, \ \hat{e}_{i}^h \ ) + (1 - \lambda) \sum\nolimits_{i=1}^N  r( A_{i}^s,\ \hat{e}_{i}^g \ ) \ . 
	\label{eq:loss_expl_model}
\end{align}
Depending on the task, the regularization function, $r(\cdot, \cdot)$, can be the \textit{RRR} term of Ross \textit{et al.} \cite{ross2017right} or the \textit{HINT} term of Selvaraju \textit{et al.} \cite{selvaraju2019taking} (\textit{cf.}~Appendix for details on these loss functions). The parameter
$\lambda$
controls how much the different feedback forms are taken into account. 
Finally, the explanatory loss 
is concatenated to the original task dependent loss term, \eg the \mbox{cross-entropy for a classification task.}
\begin{SCtable*}
	\small
	{\def\arraystretch{1.}\tabcolsep=2.pt
		\begin{tabular}{
				>{\columncolor[HTML]{C0C0C0}}r||ccccccc}
			\hline
			\cellcolor[HTML]{C0C0C0} &  &  &  &  &  &  &  \\
			\multirow{-2}{*}{\cellcolor[HTML]{C0C0C0}Data Set} & \multirow{-2}{*}{Size} & \multirow{-2}{*}{Classes} & \multirow{-2}{*}{\begin{tabular}[c]{@{}c@{}}Input-\\ dimensions\end{tabular}} & \multirow{-2}{*}{\begin{tabular}[c]{@{}c@{}}Multi-\\ object\end{tabular}} & \multirow{-2}{*}{\begin{tabular}[c]{@{}c@{}}Visual \\ confounder\end{tabular}} & \multirow{-2}{*}{\begin{tabular}[c]{@{}c@{}}Non-visual \\ confounder\end{tabular}} & \multirow{-2}{*}{\begin{tabular}[c]{@{}c@{}}Number of \\ rule-types\end{tabular}} \\ \hline
			\hline
			ToyColor \cite{ross2017right} & 40k & 2 & $5\times5\times3$ & \xmark & \cmark & \xmark & 1 \\ \hline
			ColorMNIST 
			\cite{kim2019learning} & 70k & 10 & $28\times28\times3$ & \xmark & \xmark & \cmark & 1 \\ \hline
			Decoy-MNIST \cite{ross2017right} & 70k & 10 & $28\times28\times3$ & \xmark & \cmark & \xmark & 1 \\ \hline
			Plant Data Set \cite{schramowski2020making} & 2.4k & 2 & $213\times213\times64$ & \xmark & \cmark & \xmark & 1 \\ \hline
			\cellcolor[HTML]{C0C0C0} &  &  &  &  &  &  &  \\
			\multirow{-2}{*}{\cellcolor[HTML]{C0C0C0}\begin{tabular}[c]{@{}r@{}}ISIC Skin Cancer\\Data Set \cite{codella2019skin, tschandl2018ham10000}\end{tabular}} & \multirow{-2}{*}{21k} & \multirow{-2}{*}{2} & \multirow{-2}{*}{$650\times450\times3$} & \multirow{-2}{*}{\xmark} & \multirow{-2}{*}{\cmark} & \multirow{-2}{*}{\xmark} & \multirow{-2}{*}{1} \\ \hline\hline
			\textbf{Our CLEVR-Hans3} & 13.5k & 3 & $320\times480\times3$ & \cmark & \cmark & \cmark & 2 \\ \hline
			\textbf{Our CLEVR-Hans7} & 31.5k & 7 & $320\times480\times3$ & \cmark & \cmark & \cmark & 4 \\ \hline
		\end{tabular}
	}
	\caption{\textbf{The complexity of CLEVR-Hans.}
		The CLEVR-Hans data sets represent confounded data sets in which the confounding factors are not separable in the original input space. Additionally, more than one conceptual rule must be applied in order to revise the model.
	}
	\label{table:conf_datasets}
\end{SCtable*}

\textbf{Reasoning Module.}
As the output of our concept embedding module represents an unordered set, whose class membership is unaltered by the order of the objects within the set, we require our reasoning module to handle such an input structure. The Set Transformer, recently proposed by Lee \textit{et al.}~\cite{lee2019set}, is a natural choice for such a task.

To generate the explanations of the Set Transformer given the symbolic representation, $\hat{z}_i \in [0,\ 1]^{D}$, we make use of the gradient-based Integrated Gradients explanation method of Sundararajan \textit{et al.}~\cite{sundararajan2017axiomatic}. Given a function \mbox{$g:\mathbb{R}^{N \times D} \rightarrow [0, 1]^{N \times C}$} the Integrated Gradients method estimates the importance of the $jth$ element from an input sample $z_i$, $z_{ij}$, for a model's prediction by integrating the gradients of $g(\cdot)$ along a straight line path from $z_{ij}$ to the $jth$ element of a baseline input, $\tilde{z} \in \mathbb{R}^{D}$, as $IG_j(z_i) := $
\begin{align}
	(z_{ij} - \tilde{z}_{j}) \times \int_{\alpha=0}^1 \frac{\delta \ g(\tilde{z} + \alpha \times (z_i - \tilde{z}))}{\delta z_{ij}} \delta \alpha \; .
\end{align}
Given the input to the Set Transformer, $\hat{z} \in [0,\ 1]^{N \times D}$, and $\tilde{z} = \boldsymbol{0}$ as a baseline input, we finally apply a zero threshold to only receive positive importance and thus have: 
\begin{align}
	\hat{e}^g_i := \sum\nolimits_{j=1}^{D} min(IG_j(\hat{z}_i), \ 0) \ . 
\end{align}

\textbf{(Slot) Attention is All You Need (for object-based explanations).}
Previous work of Yi \textit{et al.} \cite{yi2018neural} and Mao \textit{et al.} \cite{mao2019neuro} has shown an interesting approach for creating a Neuro-Symbolic concept leaner based on a Mask-RCNN \cite{he2017mask} scene parser. For our concept learner, we make use of the recent work of Locatello \textit{et al.}~\cite{locatello2020object}. Their proposed Slot Attention module allows to decompose the hidden representation of an encoding space into a set of task-dependent output vectors, called "slots". For example, the image encoding of a CNN backbone can be decomposed such that the hidden representation is separated into individual slots for each object. These decomposed slot encodings can then be used in further task-dependent steps, \eg attribute prediction of each object. Thus with Slot Attention, it is possible to create a fully differentiable object-centric representation of an entire image without the need to process each object of the scene individually in contrast to the system of \cite{yi2018neural, mao2019neuro}.

An additional important feature of the Slot Attention module for our setting is the ability to map each slot to the original input space via the attention maps. 
These attention maps are thus natural, intrinsic visual explanations of the detected objects. In contrast, with the scene parser of \cite{yi2018neural, mao2019neuro} it is not as straightforward to generate visual explanations based on the explanations of the reasoning module.
%
Consequently, using the Slot Attention module, we can formulate the dot-product attention for a sample $x_i$, as
\begin{equation} 
	B_i := \sigma \left( \frac{1}{\sqrt{D^{\prime}}} \ k(F_i)\cdot q(S_i)^T \right) \in \mathbb{R}^{P \times K} \ ,
\end{equation}
where $\sigma$ is the softmax function over the slots dimension, $k(F_i) \in \mathbb{R}^{P \times D^{\prime}}$ a linear projection of the feature maps $F_i$ of an image encoder for $x_i$, $q(S_i) \in \mathbb{R}^{K \times D^{\prime}}$ a linear projection of the slot encodings $S_i$ and $\sqrt{D^{\prime}}$ a fixed softmax temperature. 
$P$ represents the feature map dimensions, $K$ the number of slots and $D^{\prime}$ the dimension which the key and query functions map to. 

Finally, we can formulate  $E^h(h(\cdot), \ \hat{e}^g_{i})$ based on the attention maps $B_i$, and 
the symbolic explanation $\hat{e}_i^h$. Specifically, 
we only want an explanation for objects which were identified by the reasoning module as being relevant for the final prediction:
\begin{equation}\label{eq:sl_att_maps}
	\hat{e}_{i}^h
	:= \sum_{k=1}^K 
	\begin{cases}
		B_{ik},& \text{if } max(\hat{e}_{ik}^g)   
		\geq t\\
		\mathbf{0} \in \mathbb{R}^{P},              & \text{otherwise}
	\end{cases},
\end{equation}
where $t$ is a pre-defined importance threshold. 
Alternatively the user can manually select explanations 
for each object.

\textbf{Interchangeability of the Modules.}
Though both Slot-Attention and Set Transformer have strong advantages as stated above, alternatives exist. Deep Set Prediction Networks \cite{zhang2019deep}, Transformer Set Prediction Networks \cite{kosiorek2020conditional} or Mask-RCNN based models \cite{he2017mask} are viable alternatives to the Slot Attention module as concept embedding module. The generation of visual explanations within these models, \eg via gradient-based explanation methods, however, is not as straightforward.
Truly rule-based classifiers \cite{quinlan1986induction, loh2014fifty}, logic circuits \cite{LiangB19}, or probabilistic approaches \cite{Darwiche03, PoonD11, KisaBCD14, ManhaeveDKDR18}, are principally viable alternatives for the Set Transformer, though it remains preferable for this module to handle unordered sets. 

\section{The CLEVR-Hans Data Set} 

Several confounded computer vision data sets with varying properties, \eg number of classes, already exist.
Tab. \ref{table:conf_datasets} provides a brief overview of 
such data sets. We distinguish here between the number of samples, number of classes, image dimensions, and whether an image contains multiple objects. More important are whether a confounding factor is spatially separable from the relevant features, \eg the colored corner spots in Decoy-MNIST, whether the confounding factor is not visually separable, \eg the color in ColorMNIST that superimposes the actual digits, and finally, once the confounding factor has been identified, how many different conceptual rule-types must be applied in order to revise the model, \ie the corner rule for the digits in Decoy-MNIST is the same, regardless of which specific class is being considered. 

To the best of our knowledge, the confounded data sets listed in Tab.\ref{table:conf_datasets}, apart from ColorMNIST, possess spatially separable confounders. 
One can, therefore, revise a model by updating its spatial focus.
However, this is not possible if the confounding and true factors are not so easily separable in the input dimensions.

\begin{figure}[t]
	\centering
	\includegraphics[width=0.9\linewidth]{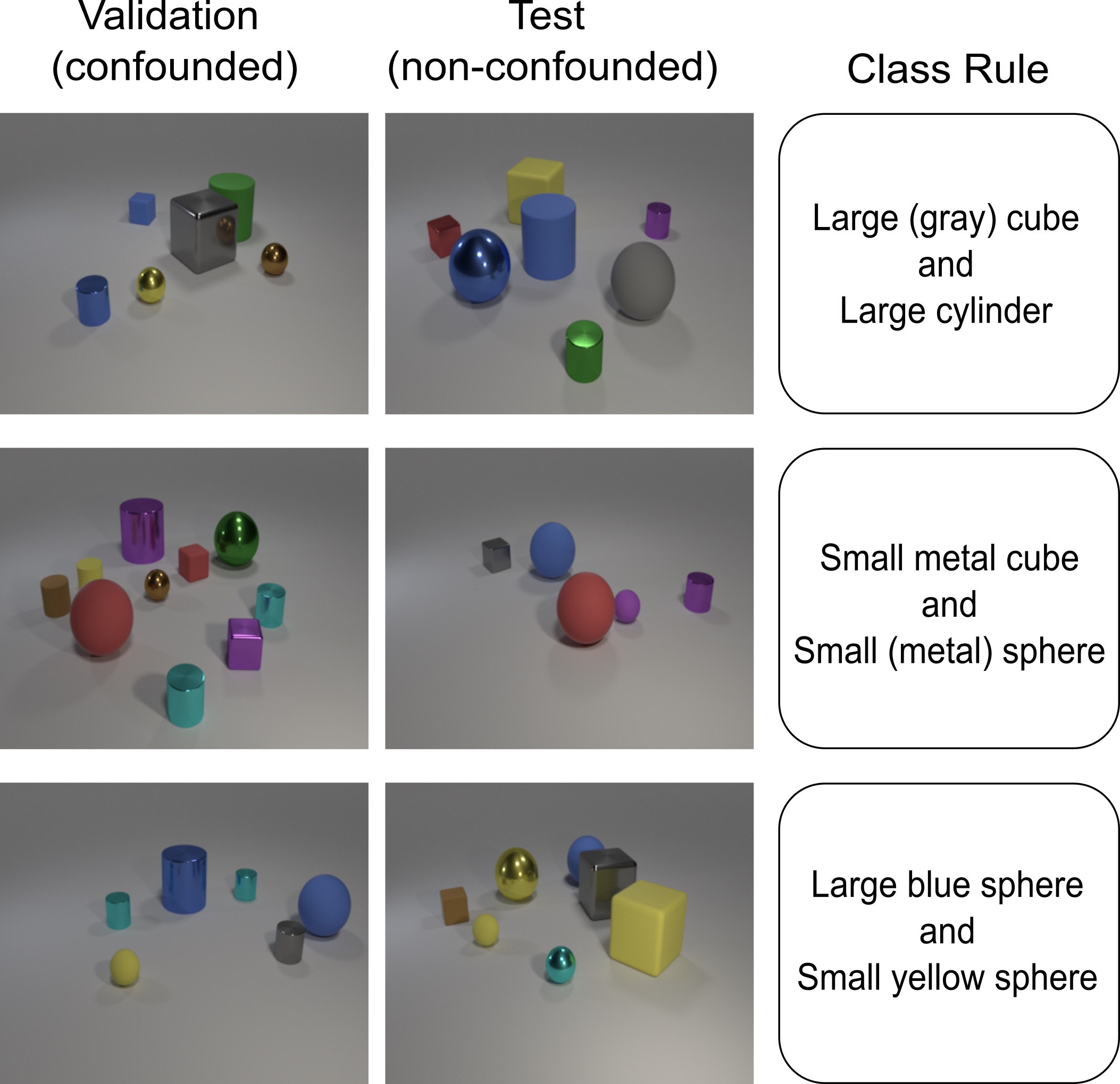}
	\caption{\textbf{Schematic of the 
			CLEVR-Hans3 data set.} Attributes in brackets are the confounding factors in the train and validation sets.}
	\label{fig:clevrhans3}
\end{figure}
The CLEVR data set of \cite{johnson2017clevr} is a particularly interesting data set, as it was originally designed to diagnose reasoning modules and presents complex scenes consisting of multiple objects and different relationships between these objects. 
Using the available framework of \cite{johnson2017clevr}, we have thus created a new confounded data set, which we refer to as the CLEVR-Hans data set. This data set consists of CLEVR images divided into several classes. The membership of a class is based on combinations of objects' attributes and relations. 
Additionally, certain classes within the data set are confounded. Thus, within the data set, consisting of train, validation, and test splits, all train, and validation images of confounded classes will be confounded with a specific attribute or combination.

We have created two variants of this data set\footnote{\url{https://github.com/ml-research/CLEVR-Hans}}, which we refer to as CLEVR-Hans3 and CLEVR-Hans7. CLEVR-Hans3 contains three classes, of which two are confounded. Fig.~\ref{fig:clevrhans3} shows a schematic representation of this data set. Images of the first class contain a large cube and large cylinder. The large cube has the color gray in every image of the train and validation set. Within the test set, the color of the large cube is shuffled randomly. Images of the second class contain a small sphere and small metal cube. The small sphere is made of metal in all training and validation set images, however, can be made of either rubber or metal in the test set. Images of the third class contain a large blue sphere and a small yellow sphere in all images of the data set. This class is not confounded.
CLEVR-Hans7 contains seven classes, of which four are confounded. 
This data set, next to containing more class rules, also contains more complex class rules than CLEVR-Hans3, \eg class rules are also based on object positions. \revision{Each class in both data sets consists of 3000 training, 750 validation, and 750 test images.} 

%
Finally, the images were created such that the exact combinations of the class rules did not occur in images of other classes. It is possible that a subset of objects from one class rule occur in an image of another class. However, it is not possible that more than one complete class rule is contained in an image. 
In summary, these data sets present an opportunity to investigate confounders and model decisions for complex 
classification rules within a benchmark data set that is more complex than previously established confounded data sets (see Tab.~\ref{table:conf_datasets}).

\section{Experimental Evidence}
Our intention here is to investigate the benefits of Neuro-Symbolic Explanatory Interactive Learning. To this end, we make use of our CLEVR-Hans data sets to investigate (1) the downsides of deep learning (DL) models in combination with current (visual) XAI methods and, in comparison, (2) the advantages of our NeSy XIL approach. In particular, we intend to investigate the benefits of  
neuro-symbolic explanations
to not just provide more detailed insights of the learned concept,
but allow for better interaction between human users and the model's explanations. We present qualitative as well as quantitative results for each experiment. \textit{Cf.}~Appendix for further details on the experiments and implementation, and additional qualitative results. 

\textbf{Architectures.} We compared our Neuro-Symbolic architecture to a ResNet-based CNN model \cite{he2016deep}, which we denote as CNN. For creating explanations of the CNN, we used the Grad-CAM method of Selvaraju \textit{et al.} \cite{selvaraju2017grad}, a back-propagation based explanation method that visualizes 
the gradients of the last hidden layer of the network's encoder, and represents a trade-off between high visual representation and spatial information. 

Due to the modular structure of our Neuro-Symbolic concept learner, Clever-Hans behavior can be due to errors within its sub-modules. As previous work \cite{ross2017right, selvaraju2019taking, teso2019explanatory, schramowski2020making} has already shown how to revise visual explanations, we did not focus on revising the visual explanations of the concept learner for our experiments. Instead, we assumed the 
concept embedding module to produce near-perfect predictions and visual explanations and focused on revising the higher-level explanations of the reasoning module. Therefore, we employed a Slot-Attention module pre-trained \revision{supervisedly} on the original CLEVR data set \cite{locatello2020object}.

\textbf{Preprocessing.} We used the same pre-processing steps as the authors of the Slot-Attention module \cite{locatello2020object}.

\textbf{Training Settings.} We trained the two models using two settings: A standard classification setting using the cross-entropy loss (Default) and the XIL setting where the explanatory loss term 
(Eq.~\ref{eq:loss_expl_model}) was appended to the cross-entropy term. The exact 
loss terms used will be discussed in the corresponding subsections.

\textbf{User Feedback.} As in \cite{teso2019explanatory, selvaraju2019taking, schramowski2020making}, we simulated the user feedback. The exact details for each experiment can be found in the corresponding subsections.

\textbf{Evaluation.} Apart from qualitatively investigating the explanations of the models, we used the classification accuracy 
on the validation and test set
as an indication of a model's ability to make predictions based on correct reasons. If the accuracy is high on the confounded validation set but low on the non-confounded test set, it is fair to assume that the model focuses on the confounding factors of the data set to achieve a high validation accuracy. 



\subsection{Visual XIL fails on CLEVR-Hans}

\begin{figure}[t]
	\centering
	\includegraphics[width=0.95\linewidth]{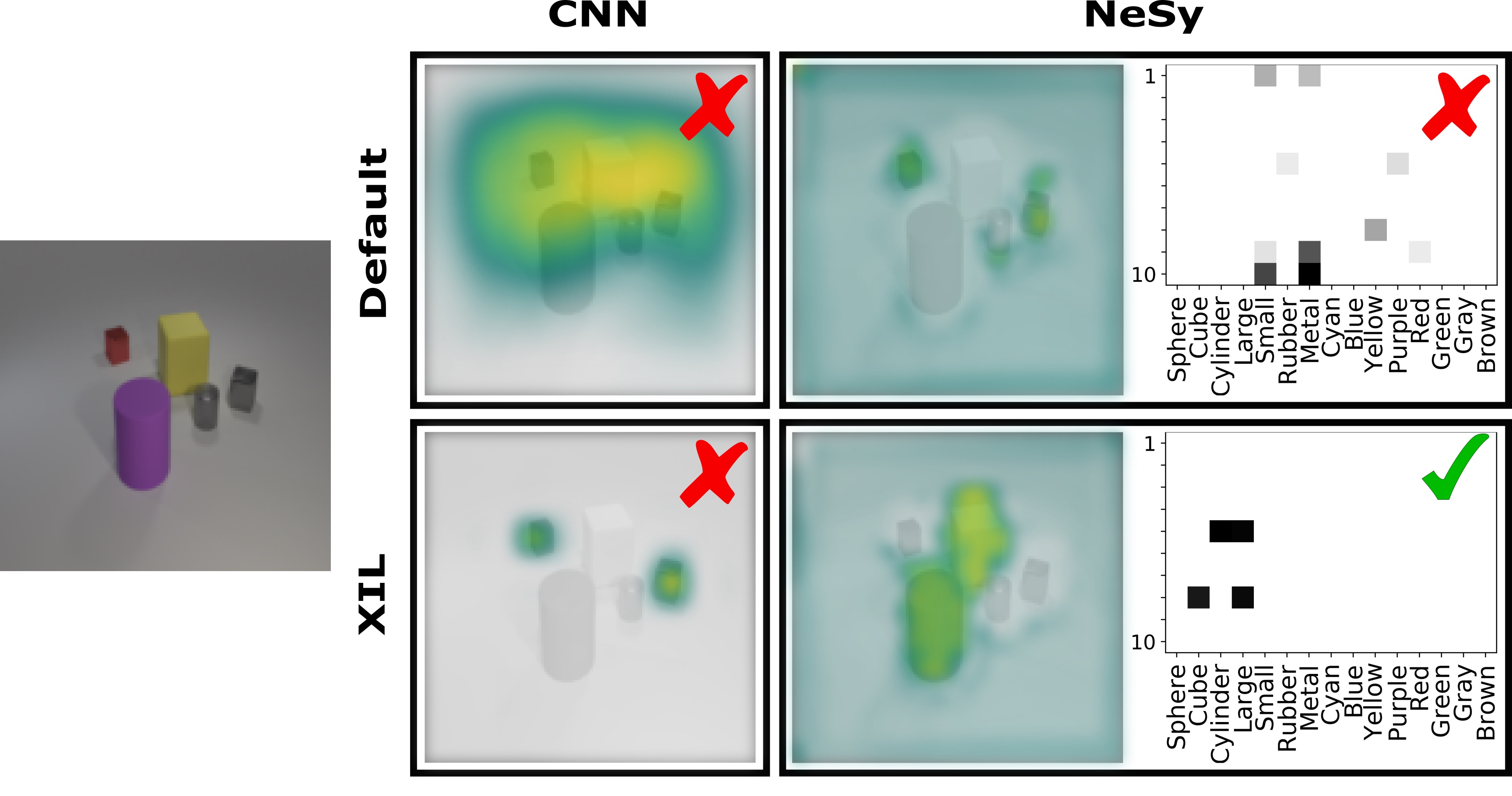}
	\caption{\textbf{Example explanations (from test set) of different model and training settings on CLEVR-Hans3.} Red crosses denote false, green checks correct predictions.}
	\label{fig:explanations}
\end{figure}

We first demonstrate the results of training a standard 
CNN for classification.

\textbf{CNN produces Clever-Hans moment.}
As Tab.~\ref{table:accs} indicates, the default CNN is prone to focusing on the confounding factors of the data sets. It reaches near perfect classification accuracies in the confounded validation sets but much lower accuracy in the non-confounded test sets. Interestingly, the main difficulty of the standard CNN for CLEVR-Hans3 appears to lie in the gray color confounder of class 1, whereas the confounding material of class $2$ does not appear to be a difficulty for the model (\textit{cf.}~Appendix).

Examples of visual explanations of the default CNN for CLEVR-Hans3 images are presented in 
Fig.~\ref{fig:explanations}. 
Note these explanations appear rather unspecific and ambiguous, and it is not clear whether the model has learned the two object class rules of CLEVR-Hans3. 

\begin{table}[t]
	\centering
	\small
	{\def\arraystretch{1.}\tabcolsep=7pt
		\begin{tabular}{r||r|r}
			\hline
			\multirow{2}{*}{\textbf{Model}} & \multicolumn{1}{c|}{\multirow{2}{*}{\begin{tabular}[c]{c}\textbf{Validation}\\(confounded)\end{tabular}}} & \multicolumn{1}{c}{\multirow{2}{*}{\begin{tabular}[c]{@{}c@{}}\textbf{Test}\\(non-confounded)\end{tabular}}} \\
			&  &  \\ \hline \hline
			\multicolumn{1}{c}{}&\multicolumn{1}{c}{\textbf{CLEVR-Hans3}}&\multicolumn{1}{c}{}\\ 
			CNN (Default) & $99.55 \pm 0.10$ & $ 70.34 \pm 0.30$ \\ \hline
			CNN (XIL) & $99.69 \pm 0.08$ & $70.77 \pm 0.37$ \\ \hline 
			NeSy (Default) & $98.55 \pm 0.27$ & $\mathbf{\circ \ 81.71 \pm 3.09}$ \\ \hline
			\textbf{NeSy XIL} & $100.00 \pm 0.00$ & $\mathbf{\bullet \ 91.31 \pm 3.13}$ \\ \hline \hline
			\multicolumn{1}{c}{}&\multicolumn{1}{c}{\textbf{CLEVR-Hans7}}&\multicolumn{1}{c}{}\\
			CNN (Default) & $96.09 \pm 0.19$ & $84.50 \pm 1.04$ \\ \hline
			CNN (XIL) & $96.08 \pm 0.25$ & $89.26 \pm 0.29$ \\ \hline 
			NeSy (Default) & $96.88 \pm 0.16$ & $\mathbf{\circ \ 90.97 \pm 0.91}$ \\ \hline
			\textbf{NeSy XIL} & $98.76 \pm 0.17$ & $\mathbf{\bullet \ 94.96 \pm 0.49}$ \\ \hline \hline
			\multicolumn{1}{c}{}&\multicolumn{1}{c}{}&\multicolumn{1}{c}{}\\
			\multicolumn{3}{c}{\textbf{CLEVR-Hans3 -- Global Correction Rule ($\boldsymbol{\lnot} \textbf{Gray}$)}}\\
			\multirow{2}{*}{\textbf{Model}} & \multicolumn{1}{c|}{\multirow{2}{*}{\begin{tabular}[c]{c}\textbf{Test}\\(class $1$)\end{tabular}}} & \multicolumn{1}{c}{\multirow{2}{*}{\begin{tabular}[c]{c}\textbf{Test}\\(all classes)\end{tabular}}} \\
			&  &  \\ \hline \hline
			NeSy (Default) & $52.98 \pm 9.60$ & $81.71 \pm 3.09$ \\ \hline
			\textbf{NeSy XIL} & $\mathbf{83.59 \pm 8.44}$ & $\mathbf{ 83.26 \pm 6.46}$ \\ \hline
			
		\end{tabular}
	}
	\caption{\textbf{Balanced accuracies on 
			Clevr-Hans3 and Clevr-Hans7}. The best (``$\bullet$'') and runner-up (``$\circ$'') results 
		are bold. We compare the test accuracy in comparison to the validation accuracy as an indication of Clever-Hans moments. 
	}
	\label{table:accs}
\end{table}
\textbf{Revising Visual Explanations via XIL.}
We next apply XIL to the CNN model to improve its explanations. As in \cite{selvaraju2019taking, schramowski2020making} we set $r(A^v, \hat{e}^v)$ to the mean squared error between user annotation and model explanation.
We simulate a user by providing ground-truth segmentation masks for each class relevant object in the train set. In this way, we could improve the model's explanations to focus more on the relevant objects of the scene. 

An example of the revised visual explanations of the CNN with XIL 
can be found in Fig.~\ref{fig:explanations} 
again visualized via Grad-CAMs. 
Compared to the not revised model, one can now clearly detect which objects are relevant for the model's prediction. However, the model's learned concept seems to not agree with the correct class rule, \textit{cf.}~Fig.~\ref{fig:clevrhans3}, and thus, in this case, it is not able to predict the correct class. Further, it is still ambiguous what concepts about those objects are relevant for the model. 
The accuracies in Tab.~\ref{table:accs} lastly indicate that correcting the visual explanations improved the overall test accuracy \revision{marginally}, however comparing to the near-perfect validation accuracy, it is clear the model still focuses on confounding factors. 

\subsection{Neuro-Symbolic XIL to the Rescue}
Now, we are ready to investigate how Neuro-Symbolic XIL improves upon visual XIL.

\textbf{Receiving Explanations of Neuro-Symbolic model.}
Training the Neuro-Symbolic model in the default cross-entropy setting, we make two observations. Firstly, we can observe an increased test accuracy compared to the previous standard CNN settings. This is likely due to the class rules' relevant features now being more evident for the model to use than the standard CNN could possibly catch on to, \eg the object's material. Secondly, even with a higher test accuracy than the previous model could achieve, this accuracy is still considerably below the again near perfect validation accuracy. This indicates that also the Neuro-Symbolic model is not resilient against confounding factors. 

Example explanations of the Neuro-Symbolic model can be found in 
Fig.~\ref{fig:explanations}, 
with the symbolic explanation on the right side and the corresponding attention-based visual explanation left of this. The objects highlighted by the visual explanations depict those objects that are considered as most relevant according to the symbolic explanation (see Eq.~\ref{eq:sl_att_maps} for details). These visualizations support the observation that the model also focuses on confounding factors.

\textbf{Revising Neuro-Symbolic Models via Interacting with Their Explanations.} We observe that the Clever-Hans moment of the Neuro-Symbolic model in the previous experiment was mainly due to errors of the reasoning module as the visual explanation correctly depicts the objects that were considered as relevant by the reasoning module. To revise the model we therefore applied XIL to the symbolic explanations via the previously used, mean-squared error 
regularization term. 
We provided the true class rules 
as semantic user feedback.

The resulting accuracies of the revised Neuro-Symbolic model can be found in Tab.~\ref{table:accs} and example explanations in Fig.~\ref{fig:explanations}. 
We observe that false behaviors based on confounding factors could largely be corrected. The XIL revised Neuro-Symbolic model produces test accuracies much higher than was previously possible in all other settings, including the XIL revised CNN.
To test the influence of possible Slot-Attention prediction errors we also tested revising the reasoning module when given the ground-truth symbolic representations. 
Indeed this way, the model could reach a near-perfect test accuracy 
(\textit{cf.}~Appendix).

\textbf{Quantitative Analysis of Symbolic Explanations.}
\revision{In order to quantitatively evaluate the symbolic explanations we compute the relative L1 error on the test set between ground-truth explanations and model explanations. Briefly, for CLEVR-Hans3 NeSy XIL resulted in a reduction in L1 error compared to NeSy (Default) of: $73\%$ (total), $64\%$ (class 1), $76\%$ (class 2) and $82\%$ (class 3). For a detailed discussion \textit{cf.}~Appendix.}

\textbf{Revision via General Feedback Rules.}
Using XIL for revising a model's explanations requires that a human user interacts with the model on a sample-based level, \ie the user receives a model's explanation for an individual sample and decides whether the explanation for this is acceptable or 
a correction on the model's explanation is necessary. This can be very tedious if a correction is not generalizable to multiple samples and must be created for each sample individually.

Consider class 1 of CLEVR-Hans3, where the confounding factor is the color gray of the large cube. Once gray has been identified as an irrelevant factor for this, but also all other classes, using NeSy XIL, a user can create a global correction rule as in Fig.~\ref{fig:overview}.
In other words, irrespective of the class label of a sample, the color gray should never play a role for prediction.
Tab.~\ref{table:accs}(bottom) shows the test accuracies of our neuro-symbolic architecture for class 1 and, separately, over all classes. We here compare the default training mode vs. XIL with the single global correction rule. For this experiment, our explanatory loss was the 
RRR term~\cite{ross2017right}, 
which has the advantage of handling negative user feedback.

As one can see, applying the correction rule has substantial advantages for class 1 test accuracies and minor advantages for the full test accuracy.
These results highlight the benefit of NeSy XIL for correcting possible Clever-Hans moments via global correction rules, a previously non-trivial feature. 

\section{Conclusion}
Neuro-Symbolic concept learners are capable of learning visual concepts by jointly understanding vision and symbolic language. However, although they combine system 1 and system 2 \cite{kahneman2011thinking} characteristics, their complexity still makes them difficult to trust in critical applications, especially, as we have shown, if the training conditions for their system 1 component may
differ from those in the test condition. However, their system 2 component allows one to identify when models are right for the wrong conceptual reasons. This allowed us to introduce the first Neuro-Symbolic Explanatory Interactive Learning approach, regularizing
a model by examining and selectively
penalizing its Neuro-Symbolic explanations. 
Our results on a newly compiled confounded benchmark data set, called CLEVR-Hans, demonstrated that semantic explanations, \textit{i.e.}, compositional explanations at a per-object, symbolic level, can identify confounders that are not identifiable using “visual” explanations only. More importantly, feedback on this semantic level makes it possible to revise the model from focusing on these confounding factors.


Our results show that Neuro-Symbolic explanations and interactions merit further investigation. Using a semantic loss \cite{xu2018} would allow one to stay at the conceptual level directly. Furthermore, one should integrate
a neural semantic parsing system that helps to interactively learn a joint symbolic language between the machine and the human user through decomposition \cite{karamcheti2020}. Lastly, language-guided XIL \cite{muetal2020shaping} is an interesting approach for more natural supervision. These approaches would help to move from XIL to conversational XIL. Applying Neuro-Symbolic prior knowledge to a model may provide additional benefits to a XIL setting. 
Finally, it is very interesting to explore more expressive reasoning components and investigate how they help combat even more complex Clever-Hans moments.
Concerning our data set, an interesting next step would be to create a confounded causal data set in the approach of \cite{GirdharR20}.

{\bf Ackowledgements.}
The authors thank the anonymous reviewers for their valuable feedback as well as Thomas Kipf for his support with Slot Attention. The work has received funding from the BMEL/BLE  under the innovation support program, project ``AuDiSens'' (FKZ28151NA187). It benefited from the Hessian research priority programme LOEWE within the
project WhiteBox as well as from the HMWK cluster project ``The Third Wave of AI.''

{\small
\bibliographystyle{ieee_fullname}
\bibliography{egbib}
}

\clearpage

\onecolumn

\begin{center}
	\section*{Appendix}
\end{center}

\subsection*{CLEVR-Hans data set}
For CLEVR-Hans classes for which class rules contain more than three objects, the number of objects to be placed per scene was randomly chosen between the minimal required number of objects for that class and ten, rather than between three and ten, as in the original CLEVR data set.

Each class is represented by 3000 training images, 750 validation images, and 750 test images. The training, validation, and test set splits contain 9000, 2250, and 2250 samples, respectively, for CLEVR-Hans3 and 21000, 5250, and 5250 samples for CLEVR-Hans7. The class distribution is balanced for all data splits. 

\begin{figure}[]
	\centering
	\includegraphics[width=0.7\linewidth]{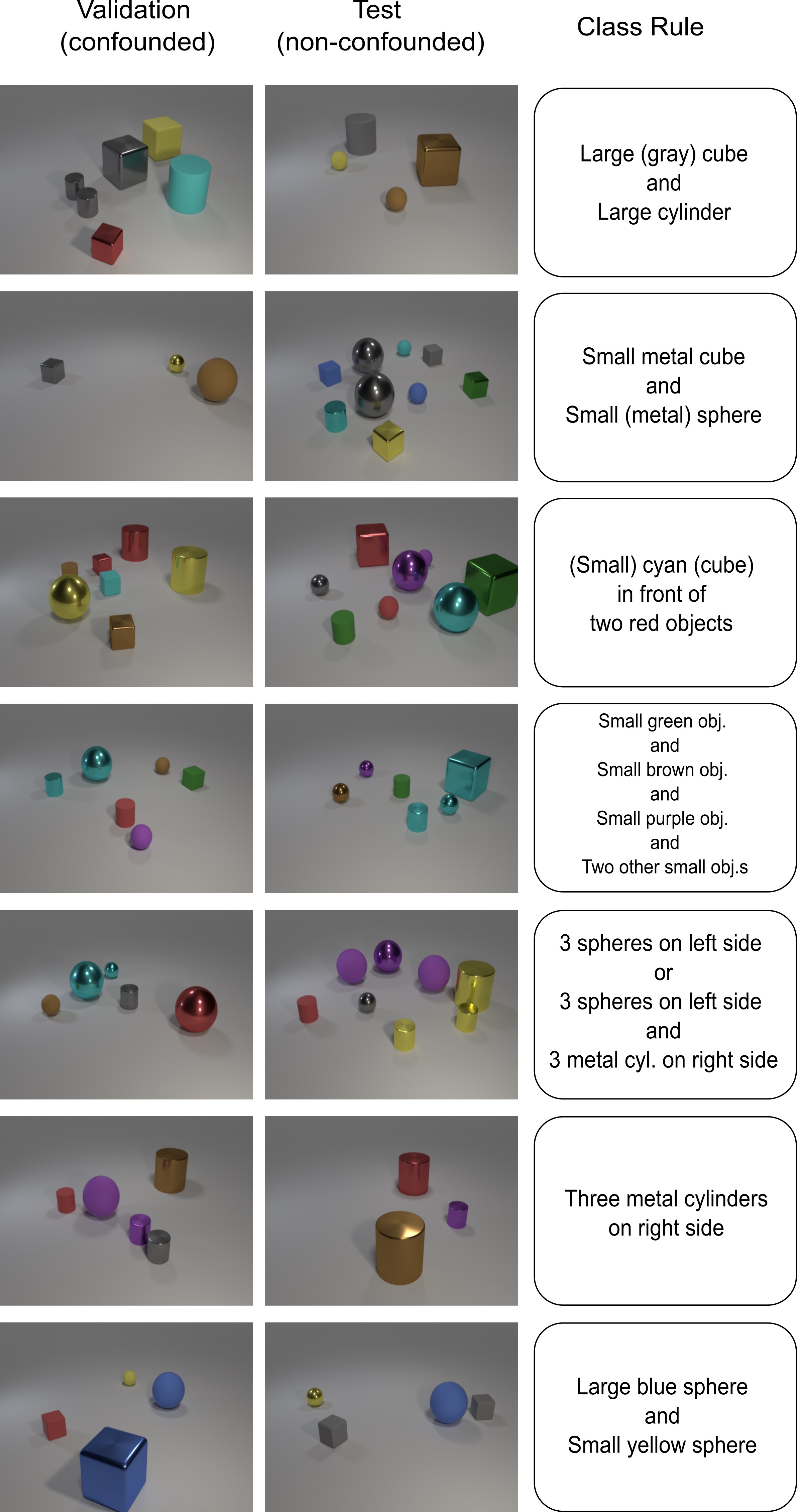}
	\caption{\textbf{CLEVR-Hans7 data set overview.} Please refer to the main text for a more detailed description of the data set.}
	\label{fig:app_clevrhans7_overview}
\end{figure}

\paragraph{CLEVR-Hans7}
The first, second, and seventh class rules of CLEVR-Hans7 correspond to classes one, two, and three of CLEVR-Hans3. Images of the third class of CLEVR-Hans7 contain a small cyan object in front of two red objects. The cyan object is a small cube in all images of the training and validation set, yet it can be any shape and size within the test set. Images of the fourth class contain at least five small objects. One of these must be green, one brown, and one purple. There are no constraints on the remaining small objects. This class is not confounded. Images of class five consist of two rules. There are three spheres present in the left half of the image (class rule 5a), or there are three spheres present in the left half of the image and three metal cylinders in the right half of the image (class rule 5b). Within all data splits, including the test split, class rule 5a occurs $90\%$ of the time and class rule 5b $10\%$ of the time. The class rule of the sixth class is contained in class rule 5b, namely three metal cylinders in the right half of the image. This is the same for all splits.

\paragraph{Preprocessing Details}
We downscaled the CLEVR-Hans images to visual dimensions 128 x 128 and normalized the images to lie between -1 and 1. For training the Slot-Attention module, an object is represented as a vector of binary values for the shape, size, color, and material attributes and continuous values between 0 and 1 for the x, y, and z positions. We refer to \cite{locatello2020object} for more details.

\subsection*{ColorMNIST Experiment}

\begin{table}[h!]
	\centering
	\begin{tabular}{cccc}
		\hline \hline
		Type                 & Size/Channels & Activation & Comment  \\ \hline \hline
		Conv 3 x 3           & 32            & ReLu       & stride 1 \\ \hline
		Conv 3 x 3           & 64            & ReLu       & stride 1 \\ \hline
		AdaptiveAvgPool (2D) & $14 \times 14$       & -          & -        \\ \hline
		Dropout              & -             & -          & p = 0.25   \\ \hline
		Flatten              & -             & -          & dim = 1    \\ \hline
		Linear               & 128           & -          & -        \\ \hline
		Dropout              & -             & -          & p = 0.5    \\ \hline
		Linear               & 10            & -          & -        \\ \hline
	\end{tabular}
	\caption{CNN used for ColorMNIST experiments.
		\label{tab:colormnist}
	}
\end{table}

The model used for the ColorMNIST data set is described in Tab~\ref{tab:colormnist}.

This model was trained with an initial learning rate of $1.0$ for 14 epochs with a batch size of $64$ using a step learning rate scheduler with step size $1$ and $\gamma = 0.7$ and Adadelta \cite{zeiler2012} as optimizer.

\subsection*{Experiment and Model Details}

\paragraph{Cross-validation}

We ran all experiments with five random parameter initializations and reported the mean classification accuracy with standard deviation over these runs. We used the seeds: 0, 1, 2, 3, 4. The accuracies we presented are from those models of each run with the lowest validation loss.

\paragraph{Reasoning Module} 

\begin{table}[]
	\centering
	\begin{tabular}{cccc}
		\hline \hline
		Type                 & Dim Out & Numb. Heads & Comment  \\ \hline \hline
		SAB           & 128            & 4       & - \\ \hline
		SAB           & 128            & 4       & - \\ \hline
		Dropout              & -             & -          & p = 0.5   \\ \hline
		PMA              & 128             & 4          & -    \\ \hline
		Dropout              & -             & -          & p = 0.5    \\ \hline
		Linear               & 3/7            & -          & -        \\ \hline
	\end{tabular}
	\caption{Set Transformer architecture used for reasoning module. Depending on whether CLEVR-Hans3 or CLEVR-Hans7 was used the final output varied between 3 and 7.
		\label{tab:settransformer}
	}
\end{table}

For our reasoning module, we used the recently proposed Set Transformer, an attention-based neural network designed to handle unordered sets. Our implementation consists of two stacked Set Attention Blocks (SAB) as encoder and a Pooling by Multihead Attention (PMA) decoder. Architecture details can be found in Tab~\ref{tab:settransformer}

\paragraph{Concept Embedding Module}  

For our concept embedding module, we used the set prediction architecture of Locatello \textit{et al.} \cite{locatello2020object} that the authors had used for the experiments on the original CLEVR data set. We refer to their paper for architecture parameters and details rather than duplicating these here.

We pre-trained this set prediction architecture on the original CLEVR data set with a cosine annealing learning rate scheduler for 2000 epochs, minimum learning rate $1e-5$, initial learning rate $4e-4$, batch size 512, 10 slots, 3 internal slot-attention iterations and the Adam optimizer \cite{KingmaB14} with $\beta_1 = 0.9$ and $\beta_2 = 0.999$,  $\epsilon = 1e-08$ and zero weight decay.

\paragraph{Neuro-Symbolic Concept Learner}

To summarize, we thus have the two modules, as stated above. For our experiments, we passed an image through the pre-trained concept embedding module. For simplicity, we binarized the output of the concept embedding module for the attributes shape, size, and color, before passing it to the reasoning module by computing the argmax of each attribute group. This way, each object is represented by a one-hot encoding of each of these attributes.

The architecture parameters of the concept embedding and reasoning module were as stated above, and the same for both training settings, i.e., default and XIL.

In the default training setting, using the cross-entropy classification loss, we used the Adam optimizer ($\beta_1 = 0.9$ and $\beta_2 = 0.999$,  $\epsilon = 1e-08$ and zero weight decay) in combination with a cosine annealing learning rate scheduler with initial learning rate $1e-4$, minimal learning rate $1e-6$, 50 epochs and batch size of 128.

For training our concept learner using the HINT \cite{selvaraju2019taking} loss term on the symbolic explanations in addition to cross entropy term we used the Adam optimizer ($\beta_1 = 0.9$ and $\beta_2 = 0.999$,  $\epsilon = 1e-08$ and zero weight decay) in combination with a cosine annealing learning rate scheduler with initial learning rate $1e-3$, minimal learning rate $1e-6$, 50 epochs and batch size of 128. We used $\lambda_s = 1000$ for the XIL experiments on CLEVR-Hans3 and $\lambda_s = 10$ for the XIL experiments on CLEVR-Hans7. For the global rule experiments, using the RRR term of Ross et al. \cite{ross2017right}, we set $\lambda_s = 20$ with all other hyperparameters the same as previously.

\paragraph{CNN Model Details} Our CNN model is based on the popular ResNet34 model of \cite{he2016deep}. The visual explanations generated by Grad-CAM are in the visual dimensions of the hidden feature maps. As these dimensions of the ResNet34 model were very coarse given our data pre-processing, we decreased the number of layers of the ResNet34 model by removing the last six convolutional layers (i.e., fourth of the four ResNet blocks) and adjusting the final linear layer accordingly. 

For training the CNN in default cross-entropy mode, we used a constant learning rate of $1e-4$ for 100 epochs and a batch size of 64. We used the Adam optimizer with $\beta_1 = 0.9$ and $\beta_2 = 0.999$,  $\epsilon = 1e-08$ and zero weight decay. For training the CNN with an additional HINT explanation regularization, we used the same training parameters, as in the default case, and a $\lambda_v = 10$. These parameters were the same for CLEVR-Hans3 and CLEVR-Hans7.

\subsection*{Explanation Loss Terms}

For our experiments, we used two different types of explanation loss terms (Eq.~4). For all experiments, apart from those with a single global rule, we simulated the user feedback as positive feedback. In other words, the user feedback indicated what features the model should be focusing on. For simplicity in our experiments, we simulated the user to have full knowledge of the task and give the fully correct rules or visual regions as feedback. For this positive feedback, we applied a simple mean-squared error between the model explanations and user feedback as an explanation loss term:
\begin{equation}
	L(\theta,\ X,\ y,\ A) = \lambda_1 \frac{1}{N} \sum_{i=1}^{N} \sum_{d=1}^{D} \left(A_{id} - \hat{e}_{id}^g\right)^2\ 
	\label{eq:mse_loss}
\end{equation}
This was applied to the XIL experiments with the standard CNN model, for which the explanations were in the form of Grad-CAMs, and for revising the Neuro-Symbolic model. In the case of revising the CNNs, the user annotation masks were downscaled to match the Grad-CAM size resulting from the last hidden layer of the CNN.

For handling the negative feedback of the experiments with the single global rule, in which the user indicated which features are not relevant, rather than which are, we reverted to the RRR term of Ross et al. \cite{ross2017right}:

\begin{equation}
	L(\theta,\ X,\ y,\ A) = \lambda_1 \sum_{i=1}^{N} \sum_{d=1}^{D} \left(A_{id} \frac{\delta}{\delta \hat{z}_{id}} \sum_{k=1}^{N_c} \log({\hat{y}}_{ik})\right)^2\ 
	\label{eq:rrr_loss}
\end{equation}

\subsection*{Quantitative Analysis of Improved Symbolic Explanations}





\begin{table}[t]
	\centering
	\small
	{\def\arraystretch{1.}\tabcolsep=7pt
		\begin{tabular}{r||r|r|r|r}
			\hline
			\multicolumn{1}{r|}{\multirow{2}{*}{\textbf{Model}}} & \multicolumn{1}{c|}{\multirow{2}{*}{\begin{tabular}[c]{@{}c@{}}\textbf{Global Test}\\ \textbf{Average}\end{tabular}}} & \multicolumn{1}{c|}{\multirow{2}{*}{\textbf{Class 1}}} & \multicolumn{1}{c|}{\multirow{2}{*}{\textbf{Class 2}}} & \multicolumn{1}{c}{\multirow{2}{*}{\textbf{Class 3}}} \\
			\multicolumn{1}{c|}{} & \multicolumn{1}{c|}{} & \multicolumn{1}{c|}{} & \multicolumn{1}{c|}{} &  \\ \hline \hline
			NeSy (Default) & $4.99 \pm 0.16$ & $6.57 \pm 0.95$ & $4.58 \pm 0.68$ & $3.81 \pm 0.34$\\ \hline
			\textbf{NeSy XIL} & $\mathbf{1.37 \pm 0.2}$ & $\mathbf{2.37 \pm 0.62}$ & $\mathbf{1.09 \pm 0.08}$ & $\mathbf{0.68 \pm 0.1}$ \\ 
			\hline \hline
			\multicolumn{1}{c}{}&\multicolumn{3}{c}{\textbf{True Positive Rate}}&\multicolumn{1}{c}{}\\ \hline
			NeSy (Default) & $2.56 \pm 0.05$ & $2.97 \pm 0.27$ & $2.13 \pm 0.23$ & $2.57 \pm 0.2$\\ \hline
			\textbf{NeSy XIL} & $\mathbf{0.85 \pm 0.09}$ & $\mathbf{1.26 \pm 0.31}$ & $\mathbf{0.77 \pm 0.06}$ & $\mathbf{0.52 \pm 0.12}$ \\ 
			\hline \hline
		\end{tabular}
	}
	\caption{\textbf{L1 error between symbolic user feedback (i.e. ground-truth (GT) symbolic explanations) and the respective model's symbolic explanations for CLEVR-Hans3.} Presented are the average L1 error over all samples of the test set and the average L1 error separately over all samples of individual classes. Note: a lower value is preferable. The best (lowest) errors are in bold. The first two rows present the L1 error over all classification errors. The bottom two rows present the error by comparing only for relevant GT elements (i.e. have a value of one).
	}
	\label{table:absdistances_clevrhans3}
\end{table}

\begin{table}[t]
	\centering
	\small
	{\def\arraystretch{1.}\tabcolsep=3pt
		\begin{tabular}{r||r|r|r|r|r|r|r|r}
			\hline
			\multirow{2}{*}{\textbf{Model}} & \multicolumn{1}{c|}{\multirow{2}{*}{\begin{tabular}[c]{c}\textbf{Global Test}\\\textbf{Average}\end{tabular}}} & \multicolumn{1}{c|}{\multirow{2}{*}{\textbf{Class 1}}} &
			\multicolumn{1}{c|}{\multirow{2}{*}{\textbf{Class 2}}} &
			\multicolumn{1}{c|}{\multirow{2}{*}{\textbf{Class 3}}} &
			\multicolumn{1}{c|}{\multirow{2}{*}{\textbf{Class 4}}} &
			\multicolumn{1}{c|}{\multirow{2}{*}{\textbf{Class 5}}} &
			\multicolumn{1}{c|}{\multirow{2}{*}{\textbf{Class 6}}} &
			\multicolumn{1}{c}{\multirow{2}{*}{\textbf{Class 7}}} \\
			\multicolumn{1}{c||}{} & \multicolumn{1}{c|}{} & \multicolumn{1}{c|}{} & \multicolumn{1}{c|}{} & \multicolumn{1}{c|}{} & \multicolumn{1}{c|}{} & \multicolumn{1}{c|}{} & \multicolumn{1}{c|}{} \\ \hline \hline
			NeSy (Default) & $7.26 \pm 0.32$ & $6.14 \pm 0.65$ & $7.61 \pm 0.66$ & $6.64 \pm 0.33$ & $6.93 \pm 1.83$ & $10.1 \pm 0.45$ & $7.66 \pm 0.73$ & $5.77 \pm 0.51$\\ \hline
			\textbf{NeSy XIL} & $\mathbf{3.35 \pm 0.13}$ & $\mathbf{2.28 \pm 0.09}$ & $\mathbf{2.86 \pm 0.08}$ & $\mathbf{4.72 \pm 0.72}$ & $\mathbf{1.88 \pm 0.27}$ & $\mathbf{7.19 \pm 0.45}$ & $\mathbf{2.90 \pm 0.24}$ & $\mathbf{1.59 \pm 0.09}$\\ 
			\hline \hline
			\multicolumn{1}{c}{}&\multicolumn{7}{c}{\textbf{True Positive Rate}}&\multicolumn{1}{c}{}\\ \hline
			NeSy (Default) & $4.55 \pm 0.12$ & $3.35 \pm 0.27$ & $4.12 \pm 0.26$ & $4.41 \pm 0.16$ & $3.13 \pm 0.55$ & $7.6 \pm 0.08$ & $5.57 \pm 0.38$ & $3.65 \pm 0.22$\\ \hline
			\textbf{NeSy XIL} & $\mathbf{2.43 \pm 0.12}$ & $\mathbf{1.38 \pm 0.04}$ & $\mathbf{1.87 \pm 0.11}$ & $\mathbf{3.22 \pm 0.51}$ & $\mathbf{1.48 \pm 0.26}$ & $\mathbf{6.03 \pm 0.5}$ & $\mathbf{2.00 \pm 0.28}$ & $\mathbf{1.04 \pm 0.06}$\\ 
			\hline \hline
			
		\end{tabular}
	}
	\caption{\textbf{L1 error between symbolic user feedback (i.e. ground-truth (GT) symbolic explanations) and the respective model's symbolic explanations for CLEVR-Hans7.} Presented are the average L1 error over all samples of the test set and the average L1 error separately over all samples of individual classes. Note: a lower value is preferable. The best (lowest) errors are in bold. The first two rows present the L1 error over all classification errors. The bottom two rows present the error by comparing only for relevant GT elements (i.e. have a value of one).
	}
	\label{table:absdistances_clevrhans7}
\end{table}

To more quantitatively assess the improvements of the symbolic explanations of our NeSy model using XIL we measured the absolute difference (L1 error) for each sample between the ground-truth (GT) explanations and the symbolic explanations of the NeSy Default trained with cross-entropy and NeSy XIL, respectively. Specifically, we computed the difference for an individual sample in the following. Given the GT explanation $e^{GT}_i \in [0,\ 1]^{D}$ and symbolic explanation of the model $\hat{e}^{g}_i \in [0,\ 1]^{D}$ of sample $i$ we computed the L1 error as: $\sum_j^D |e^{GT}_{ij} - \hat{e}^{g}_{ij} |$. We finally averaged the error over all samples of the test set, as well as all samples of a specific sample class, separately.

Due to that within $e^{GT}_i$ only few attributes are marked as relevant (i.e. have a value of one) we measured the absolute L1 error here over all possible classification errors, i.e. true positives, true negatives, false positives and false negatives. The results can be found in the top two rows of Tab.~\ref{table:absdistances_clevrhans3} and Tab.~\ref{table:absdistances_clevrhans7} for CLEVR-Hans3 and CLEVR-Hans7, respectively. Note here that a lower error corresponds to a stronger correspondence between the GT explanation and model explanation.

Additionally we computed the absolute L1 error only over the relevant GT attributes, yielding the true positive rate. The results can be found in the bottom two rows Tab.~\ref{table:absdistances_clevrhans3} and Tab.~\ref{table:absdistances_clevrhans7} for CLEVR-Hans3 and CLEVR-Hans7, respectively. One can observe that in fact with XIL the symbolic model explanations more strongly correspond to the GT explanations, thus further supporting the results indicated by the balanced accuracies for validation and test sets of the main text as well as the qualitative results of the main text and supplementary materials that using XIL on the symbolic explanations the model explanations could be improved to more strongly correspond to the underlying GT symbolic explanations.

For CLEVR-Hans7 NeSy XIL resulted in a reduction in relative L1 error compared to NeSy (Default) of: $54\%$ (total), $63\%$ (class 1), $62\%$ (class 2), $29\%$ (class 3), $73\%$ (class 4), $29\%$ (class 5), $62\%$ (class 6) and $72\%$ (class 7). 

One particularly interesting result to take from Tab.~\ref{table:absdistances_clevrhans7} is the difficulty of improving the symbolic explanations for those classes of CLEVR-Hans7 which require counting the occurrences of specific attribute combinations, i.e. classes 3 and 5 (see Fig.~\ref{fig:app_clevrhans7_overview} for an overview of the class rules). The improvement in L1 error for NeSy XIL is not as strong for class 3 and class 5 as for the other classes. We believe this to indicate a shortcoming in the counting ability of the Set Transformer module.

\subsection*{Detailed Analysis of Confusion Matrices}


\begin{figure*}
	\centering
	\begin{subfigure}[b]{0.495\textwidth}
		\centering
		\includegraphics[width=\textwidth]{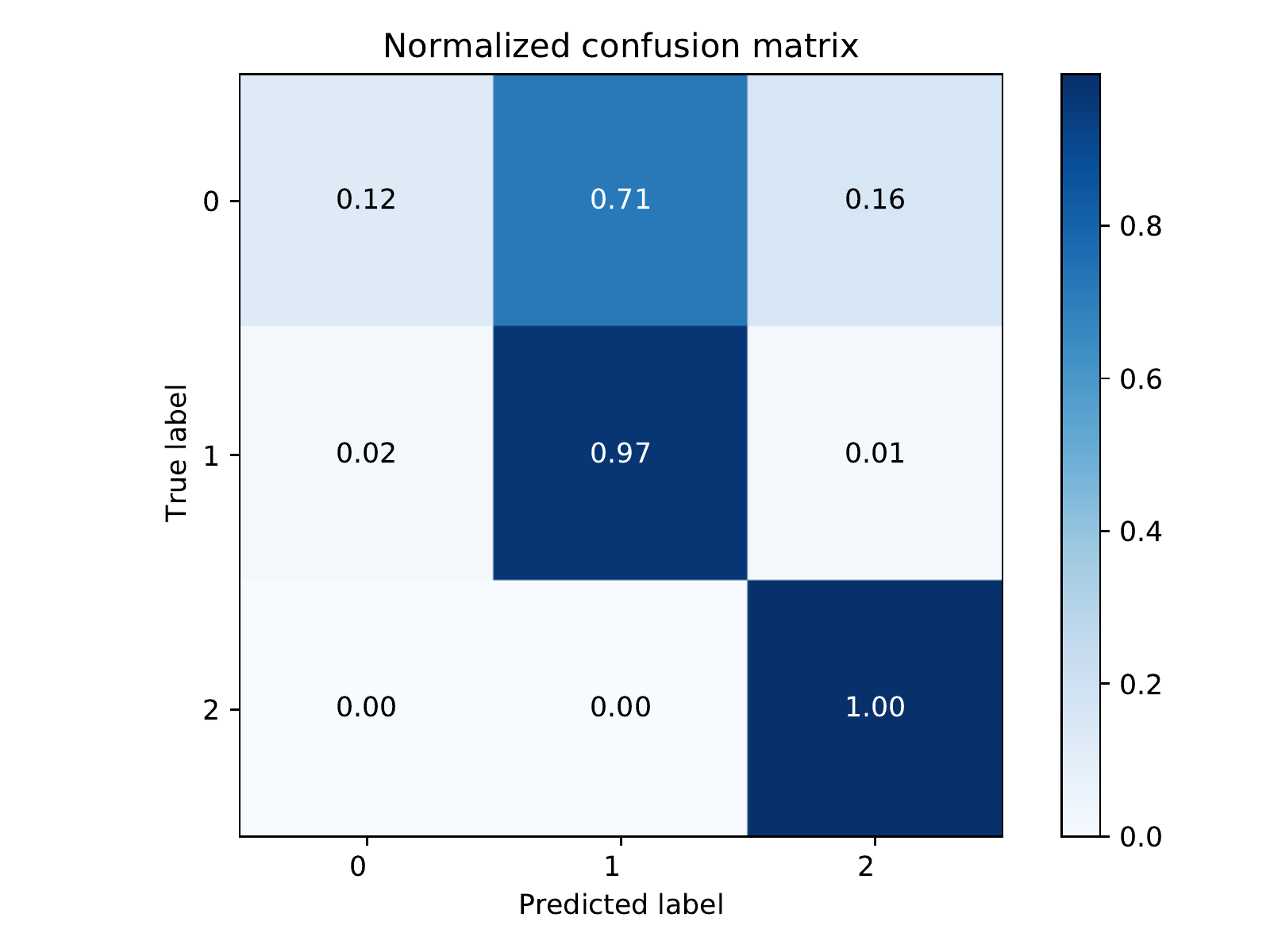}
		\caption{CNN Default}    
		\label{}
	\end{subfigure}
	\hfill
	\begin{subfigure}[b]{0.495\textwidth}  
		\centering 
		\includegraphics[width=\textwidth]{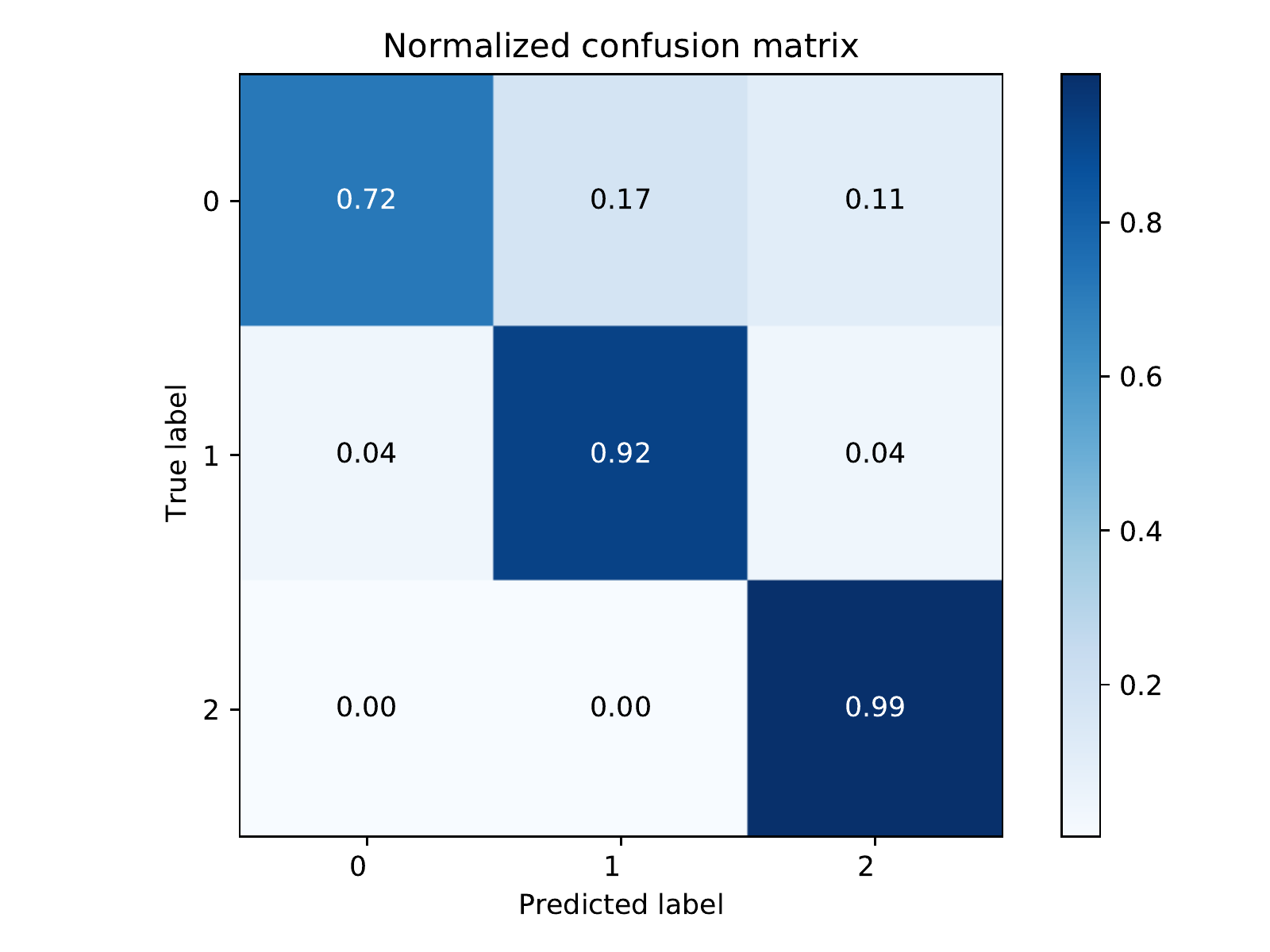}
		\caption{NeSy Default}    
		\label{}
	\end{subfigure}
	\vskip\baselineskip
	\begin{subfigure}[b]{0.495\textwidth}   
		\centering 
		\includegraphics[width=\textwidth]{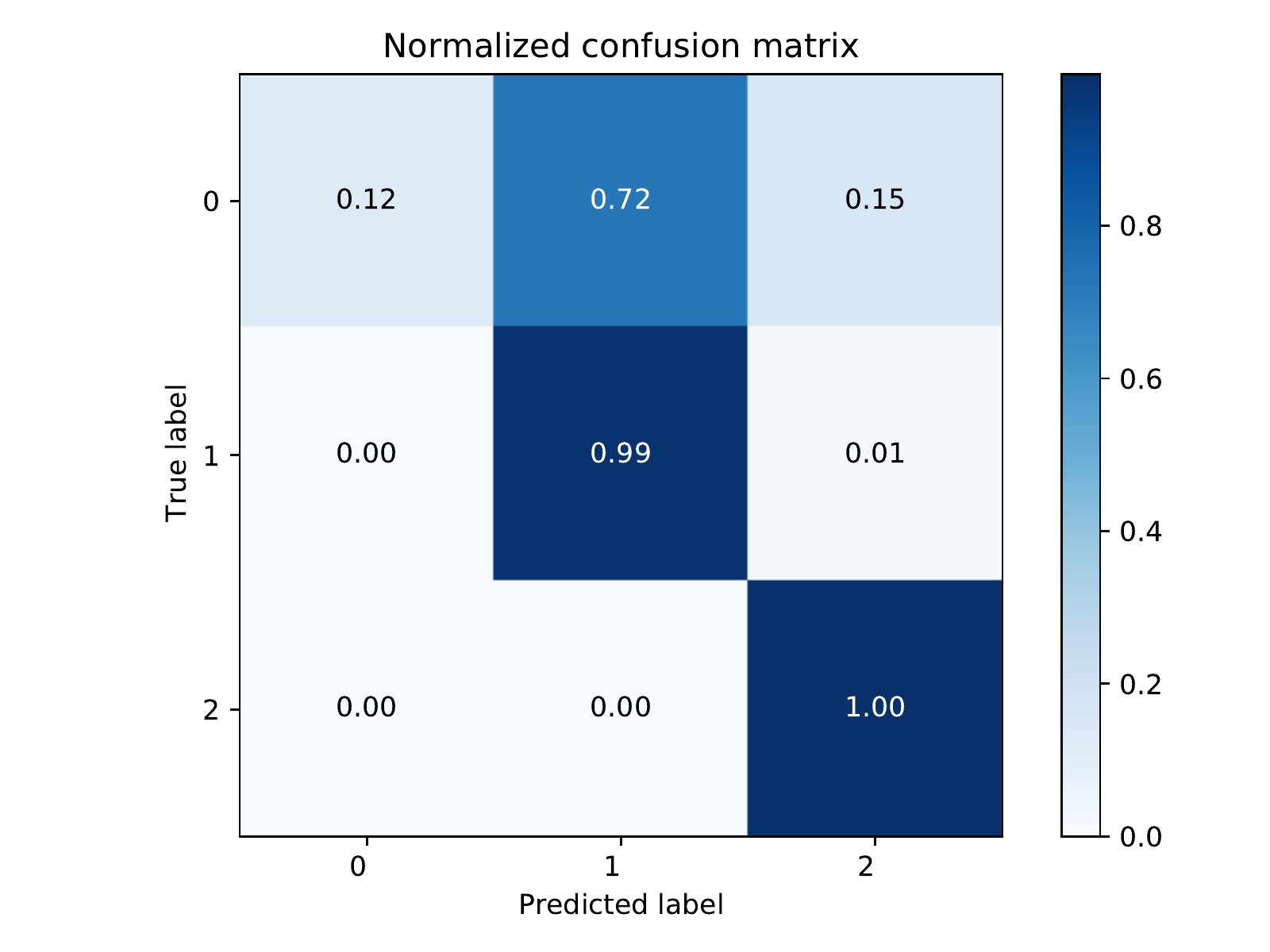}
		\caption{CNN XIL}    
		\label{}
	\end{subfigure}
	\hfill
	\begin{subfigure}[b]{0.495\textwidth}   
		\centering 
		\includegraphics[width=\textwidth]{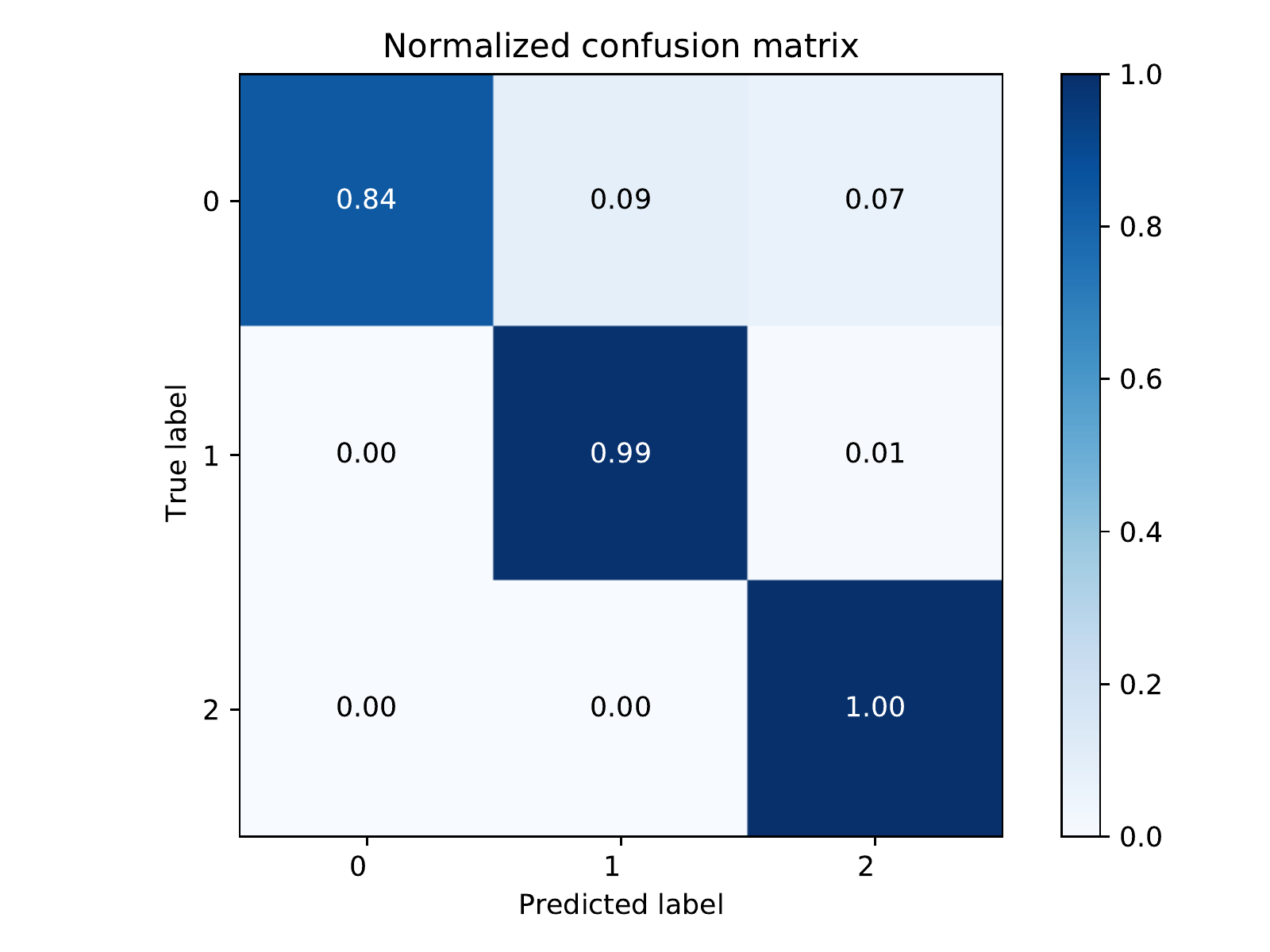}
		\caption{NeSy XIL}    
		\label{}
	\end{subfigure}
	\caption{\textbf{Confusion matrices of the different models and training settings for the test set of CLEVR-Hans3.} Note: label 0 corresponds to class 1, etc..}
	\label{fig:conf_matrix_cnn_conf3} 
\end{figure*}

\begin{figure*}
	\centering
	\begin{subfigure}[b]{0.495\textwidth}
		\centering
		\includegraphics[width=\textwidth]{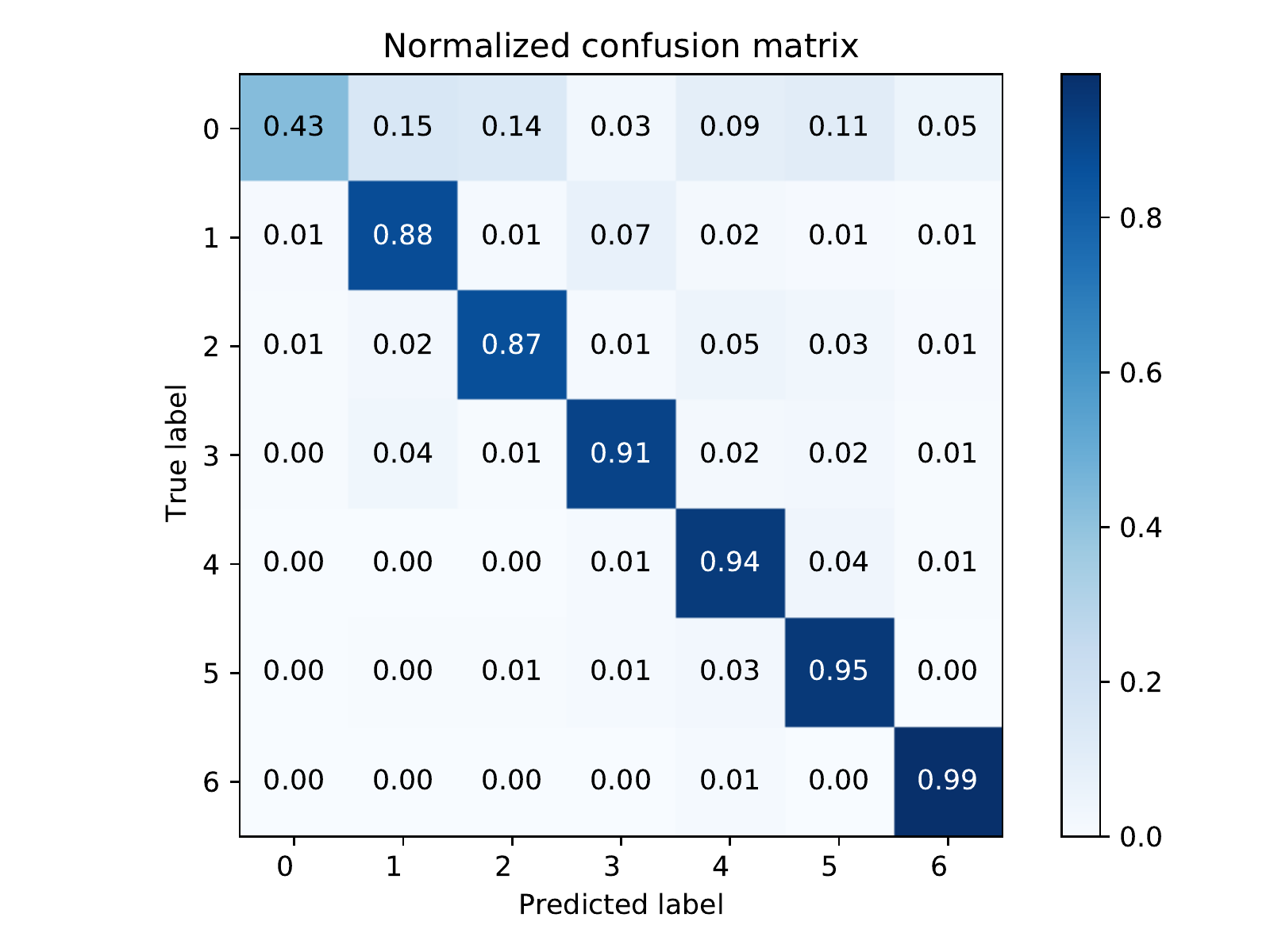}
		\caption{CNN Default}    
		\label{}
	\end{subfigure}
	\hfill
	\begin{subfigure}[b]{0.495\textwidth}  
		\centering 
		\includegraphics[width=\textwidth]{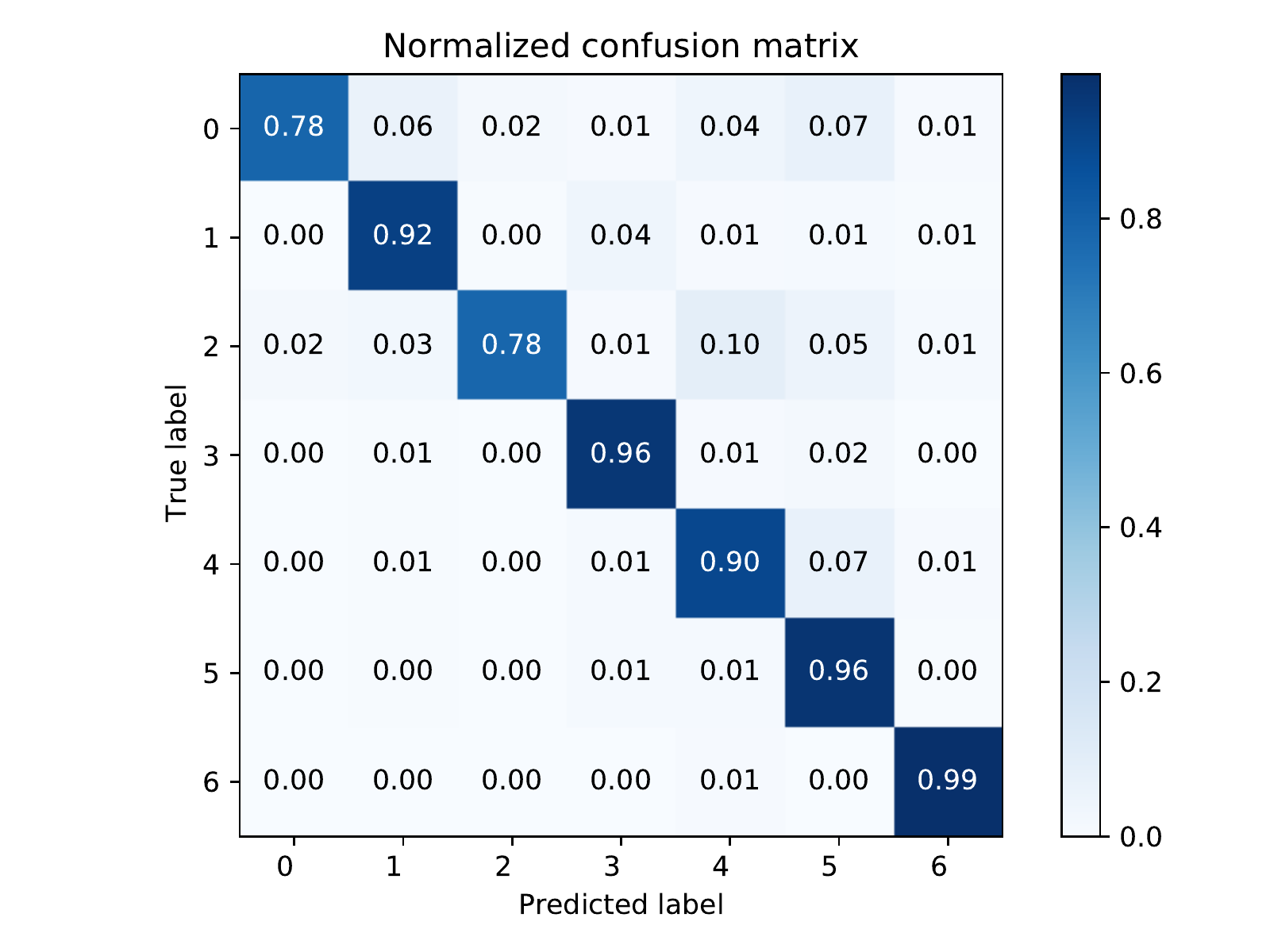}
		\caption{NeSy Default}    
		\label{}
	\end{subfigure}
	\vskip\baselineskip
	\begin{subfigure}[b]{0.495\textwidth}   
		\centering 
		\includegraphics[width=\textwidth]{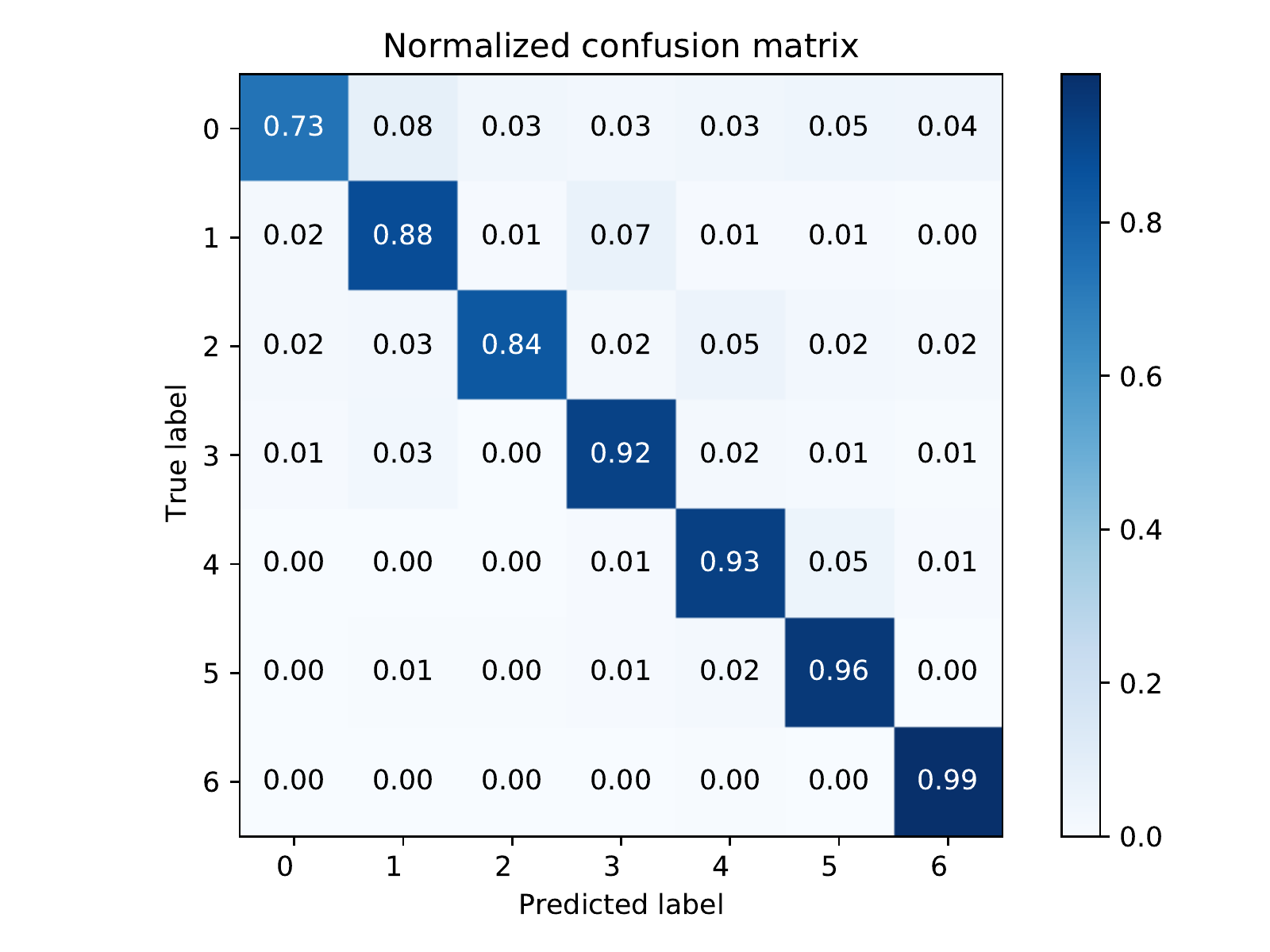}
		\caption{CNN XIL}    
		\label{}
	\end{subfigure}
	\hfill
	\begin{subfigure}[b]{0.495\textwidth}   
		\centering 
		\includegraphics[width=\textwidth]{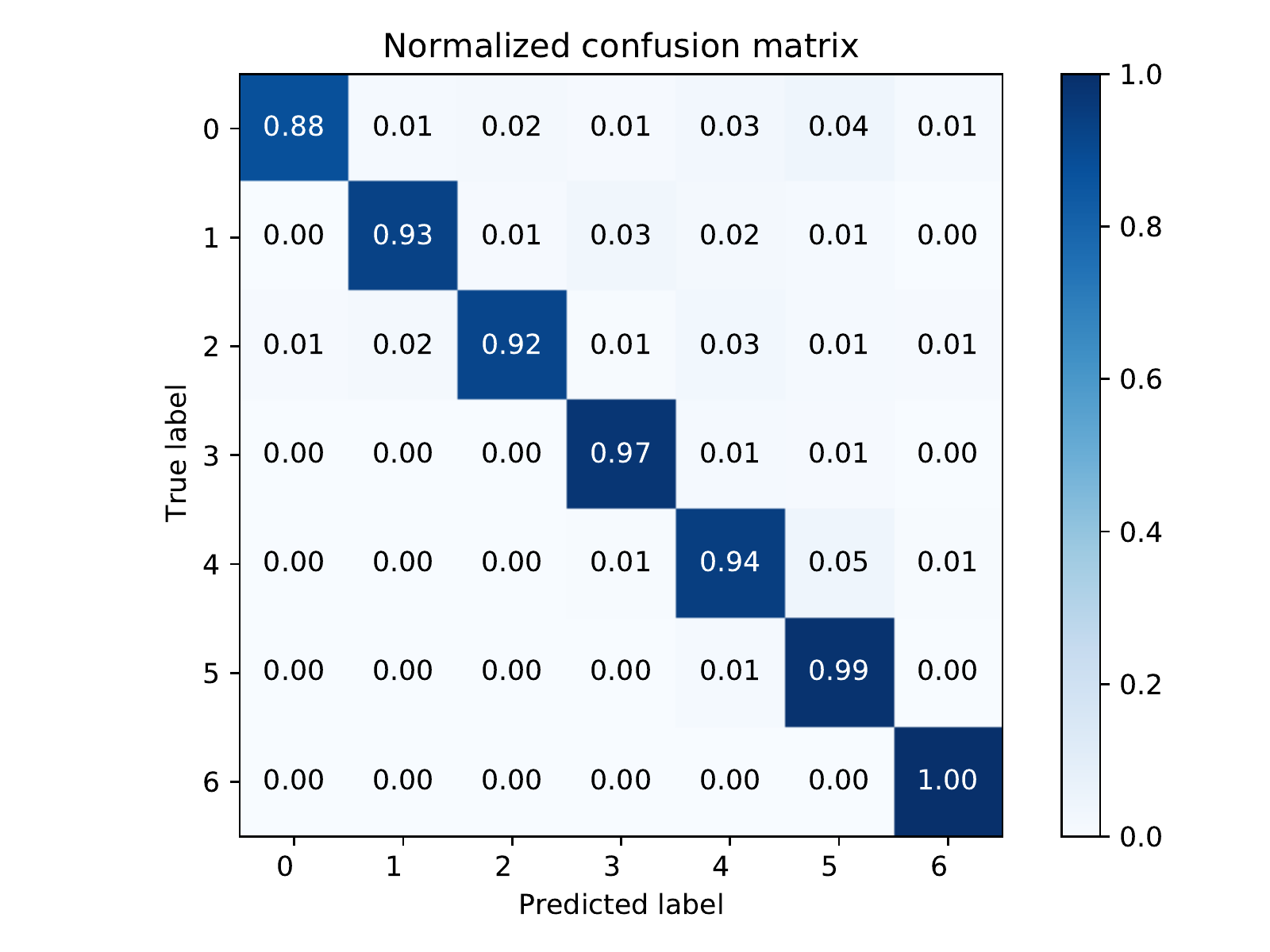}
		\caption{NeSy XIL}    
		\label{}
	\end{subfigure}
	\caption{\textbf{Confusion matrices of the different models and training settings for the test set of CLEVR-Hans3.} Note: label 0 corresponds to class 1, etc..}
	\label{fig:conf_matrix_cnn_conf7} 
\end{figure*}

Fig.~\ref{fig:conf_matrix_cnn_conf3} presents the confusion matrix for the all model and training settings on the test set of CLEVR-Hans3. Note the default CNN's difficulty especially with the color confounder of class one rather than the material confounder of class two. 

Fig.~\ref{fig:conf_matrix_cnn_conf7} presents the confusion matrix for the all model and training settings on the test set of CLEVR-Hans7. Quite surprisingly, in comparison to Fig.~\ref{fig:conf_matrix_cnn_conf3} one can see that within the classes also present in CLEVR-Hans3 all models reach a higher class accuracy than when trained with CLEVR-Hans3. We suggest this is caused by the nonexclusive nature of the CLEVR-Hans data generation. As an example: though a large gray cube and large cylinder will never appear in combination in any other image than of class 1, each object separately may appear in images of other classes. Thus with more images available in which an individual large gray cube may appear, the confounding factor, the color gray, may not carry as much weight as with fewer classes and images. Thus the generalizability to the test set is from the start easier to accomplish with CLEVR-Hans7.

\subsection*{Additional Explanation Visualizations}

\begin{figure}[]
	\centering
	\begin{subfigure}[b]{0.49\linewidth}
		\centering
		\includegraphics[width=0.9\textwidth]{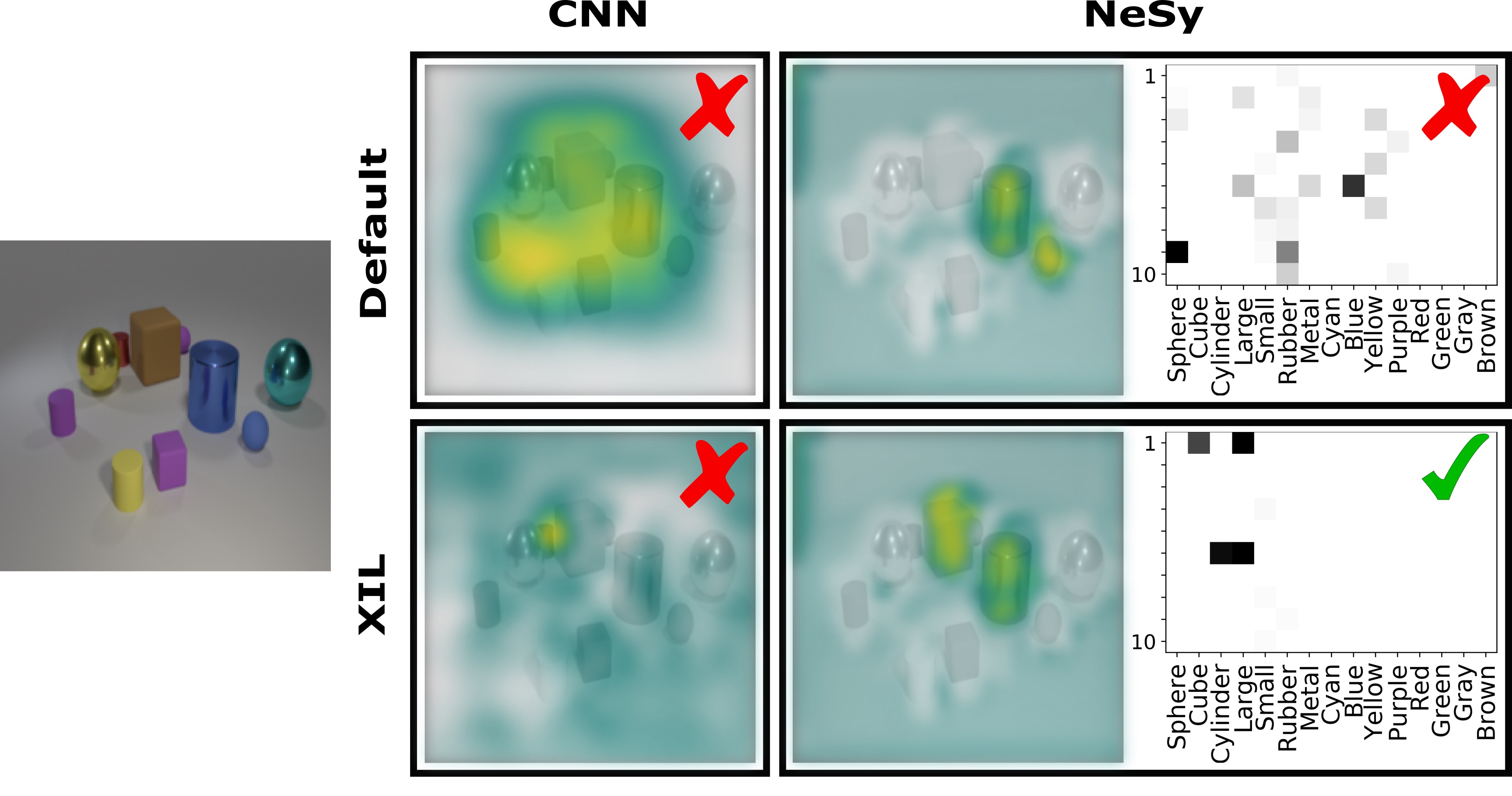}
		\caption{}
	\end{subfigure}
	\begin{subfigure}[b]{0.49\linewidth}
		\centering
		\includegraphics[width=0.9\textwidth]{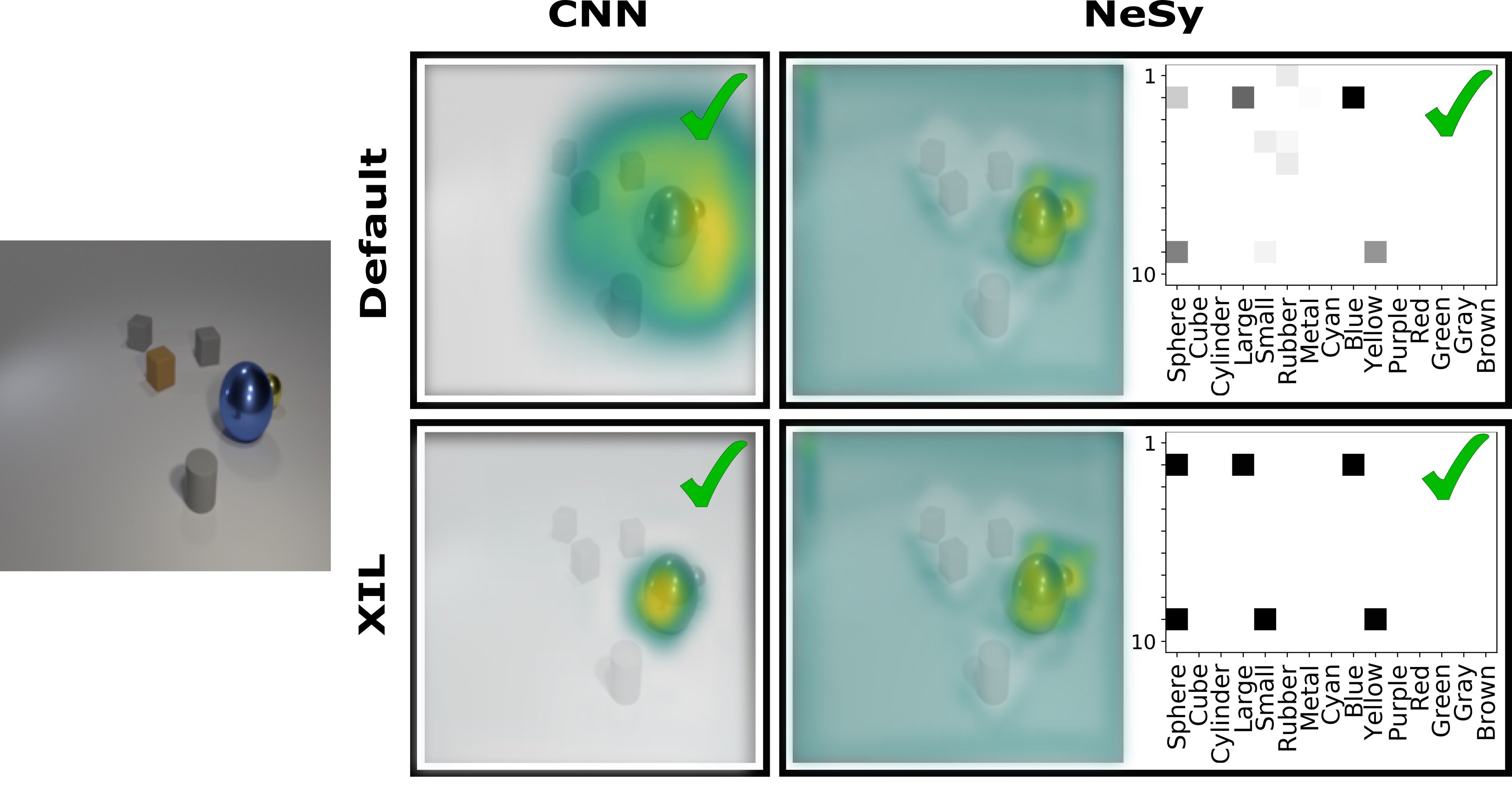}
		\caption{}
	\end{subfigure}
	\begin{subfigure}[b]{0.49\linewidth}
		\centering
		\includegraphics[width=0.9\textwidth]{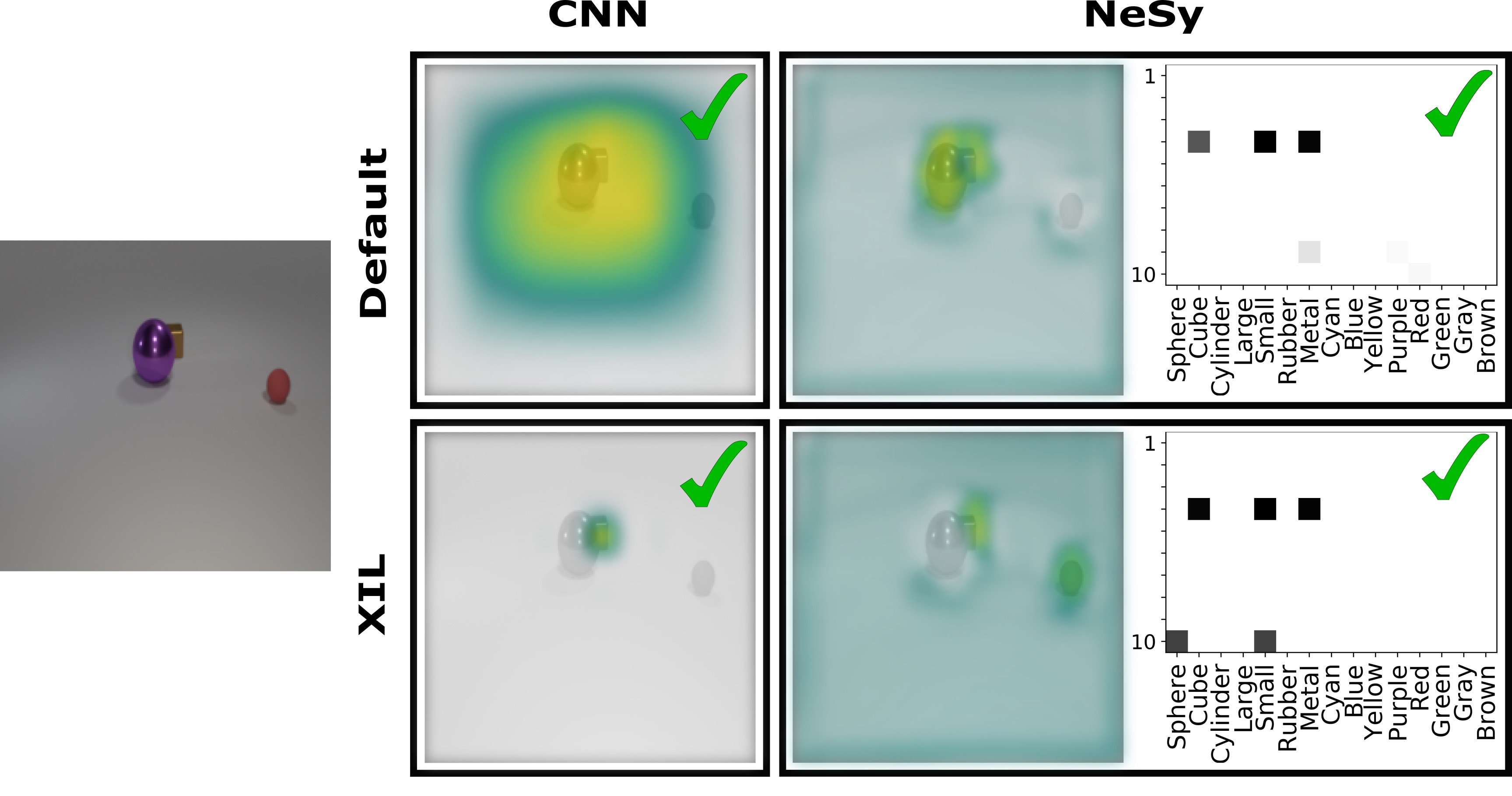}
		\caption{}
	\end{subfigure}
	\begin{subfigure}[b]{0.49\linewidth}
		\centering
		\includegraphics[width=0.9\textwidth]{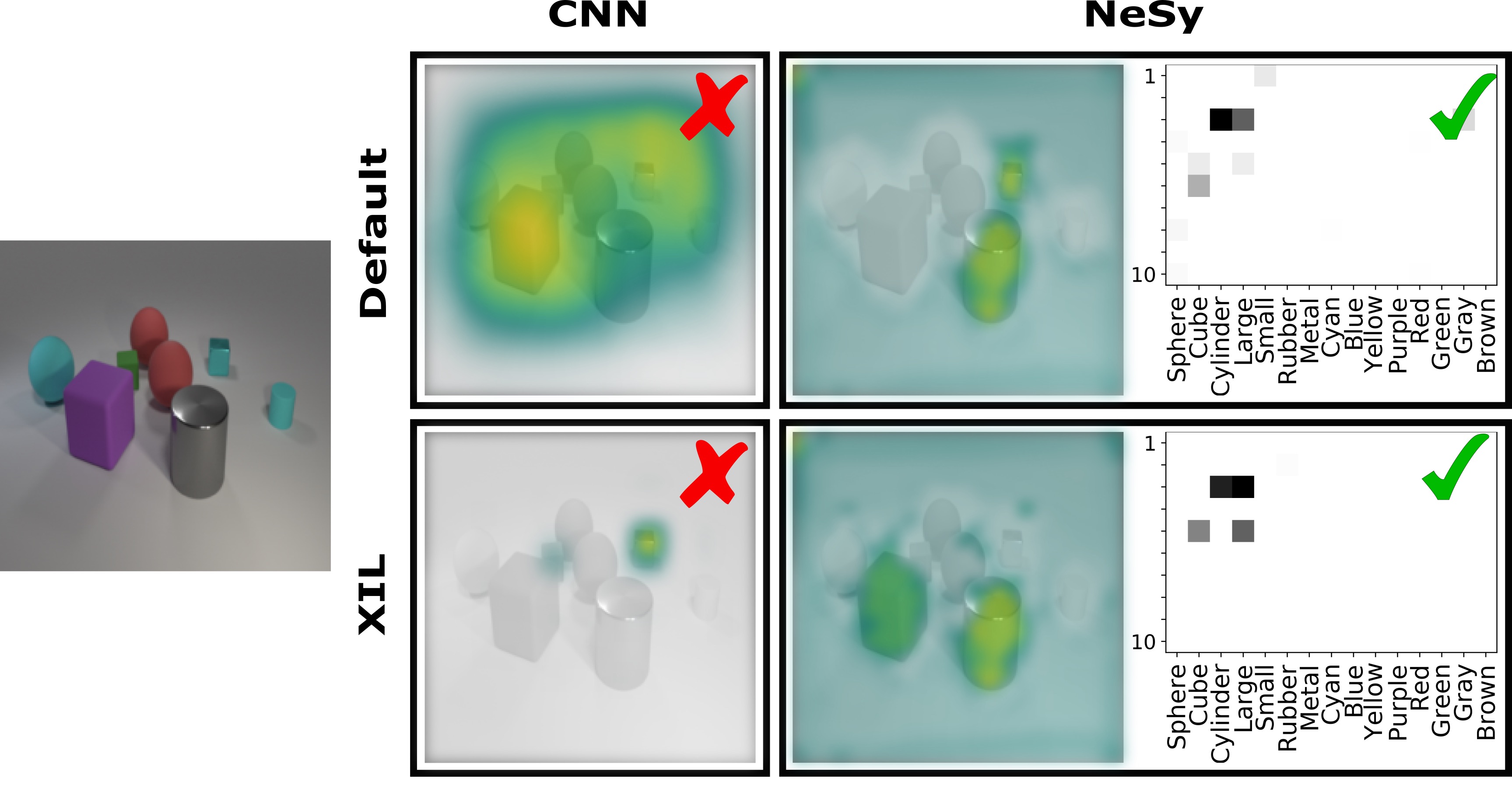}
		\caption{}
	\end{subfigure}
	\begin{subfigure}[b]{0.5\linewidth}
		\centering
		\includegraphics[width=0.9\textwidth]{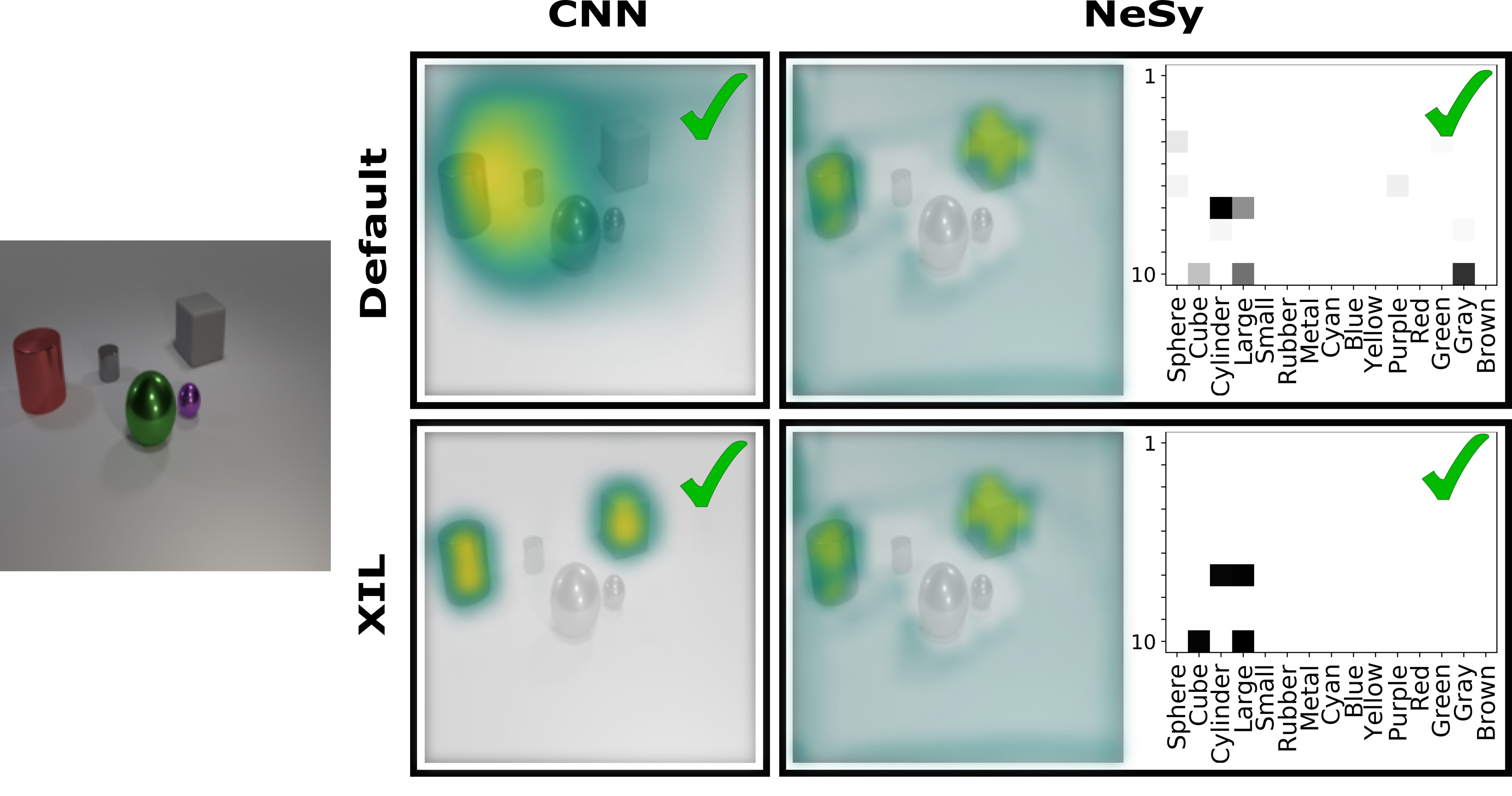}
		\caption{}
	\end{subfigure}
	\caption{Additional explanations of the various model types for test samples. Green checks represent correct class predictions, red crosses incorrect predictions.}
	\label{fig:app_expls}
\end{figure}

Fig.~\ref{fig:app_expls} shows additional qualitative results of NeSy XIL in addition to those of the main text. The top left example (a) presents another example where only via interacting with Neuro-Symbolic explanations can get the correct prediction for the correct reason. Top right (b) shows an example where all model configurations make the correct prediction. However, it does not become clear whether the CNN is indeed focusing on both relevant objects. With the NeSy model, this becomes clearer, though only using NeSy XIL are the correct objects and attributes identified as relevant for prediction. A similar case can be found in the middle left (c), where NeSy XIL aids in focusing on both relevant objects. The middle right shows a case where already NeSy shows advantages for creating correct predictions, yet not entirely for the correct concept. The bottom example (e) exemplifies that solely from a visual explanation, it does not become clear that the model is focusing on the color confounder, gray.

\subsection*{Further Concluding Remarks}

The presented CLEVR-Hans benchmarks are challenging data sets due to the complex logic concepts that underlie the visual scenes, we also strive towards an evaluation on real world data sets. Since, Koh et al. \cite{koh2020concept} and Kim et al. \cite{kim2020visual} show that the performance of concept based models on real world data sets are en par with popular black-box models ---however, don't investigate revising these models--- we expect good results here as well.

\clearpage


\end{document}